%% file: main.tex
\documentclass[journal]{IEEEtran}

\usepackage{cite}
\usepackage{xspace}
\usepackage{subcaption}
\usepackage{amsmath, bbm}
\usepackage{wrapfig}
\usepackage{captdef}
\usepackage{arydshln}
\usepackage{comment, makecell}
\usepackage{multirow}
\usepackage{graphicx}
\usepackage{enumitem}
\usepackage{algorithm}
\usepackage{algorithmic}
\usepackage{cleveref}
\usepackage{amsmath, amsthm}
\DeclareMathOperator*{\argmax}{arg\,max}
\DeclareMathOperator*{\argmin}{arg\,min}
\usepackage{booktabs}
\usepackage{array,multirow}
\usepackage{amsfonts}
\crefformat{section}{\S#2#1#3}
\crefformat{subsection}{\S#2#1#3}
\crefformat{subsubsection}{\S#2#1#3}
\usepackage{xcolor}

\newcommand{\proposed}{BPL\xspace}
\newcommand{\so}{$\mathcal{S}^1$\xspace}
\newcommand{\sz}{$\mathcal{S}^0$\xspace}

\newcommand{\szo}{$\mathcal{S}^{01}$\xspace}
\newcommand{\s}{$\mathcal{S}$\xspace}
\newcommand{\smallsection}[1]{{\vspace{0.05in} \noindent \bf {#1}}}
\usepackage{url}

\newlist{researchquestions}{enumerate}{1}
\setlist[researchquestions]{label*=\textbf{RQ\arabic*}}

\renewcommand{\textcolor}[2]{#2}
\renewcommand{\color}[1]{}

\begin{document}
\title{BPL: Bias-adaptive Preference Distillation Learning for~Recommender System}
%
%
%
\author{SeongKu Kang\IEEEauthorrefmark{1},
        Jianxun Lian,
        Dongha Lee,
        Wonbin Kweon,
        Sanghwan Jang,
        Jaehyun Lee,\\
        Jindong Wang,
        Xing Xie,~\IEEEmembership{Fellow,~IEEE,}
        and Hwanjo Yu
\thanks{SeongKu Kang is with the Department of Computer Science and Engineering, Korea University, Seoul, South Korea. E-mail:seongkukang@korea.ac.kr.}
\thanks{Jianxun Lian and Xing Xie are with Microsoft Research Asia. E-mail: \{jianxun.lian, xingx\}@microsoft.com.}
\thanks{Dongha Lee is with the Department of Aritifial Intelligence, Yonsei University, Seoul, South Korea, E-mail:donalee@yonsei.ac.kr.}
\thanks{Wonbin Kweon, Sanghwan Jang, Jaehyun Lee, and Hwanjo Yu are with the Department of Computer Science and Engineering, POSTECH, Pohang, South Korea, Email: \{kwb4453, jsh710101, jminy8, hwanjoyu\}@postech.ac.kr}
\thanks{Jindong Wang is with William \& Mary, Virginia, United States. E-mail:jwang80@wm.edu.}
\thanks{\IEEEauthorrefmark{1}Corresponding author.}
}

\maketitle

\begin{abstract}
Recommender systems suffer from biases that cause the collected feedback to incompletely reveal user preference. 
While debiasing learning has been extensively studied, they mostly focused on the specialized (called \textit{counterfactual}) test environment simulated by random exposure of items, significantly degrading accuracy in the typical (called \textit{factual}) test environment based on actual user-item interactions.
In fact, each test environment highlights the benefit of a different aspect:
the counterfactual test emphasizes user satisfaction in the long-terms, while the factual test focuses on predicting subsequent user behaviors on platforms.
Therefore, it is desirable to have a model that performs well on both tests rather than only one.
In this work, we introduce a new learning framework, called \textbf{B}ias-adaptive \textbf{P}reference distillation \textbf{L}earning (\proposed), to gradually uncover user preferences with dual distillation strategies.
These distillation strategies are designed to drive high performance in both factual and counterfactual test environments.
Employing a specialized form of \textit{teacher-student distillation} from a biased model, \proposed retains accurate preference knowledge aligned with the collected feedback, leading to high performance in the factual test.
Furthermore, through \textit{self-distillation} with reliability filtering, \proposed iteratively refines its knowledge throughout the training process.
This enables the model to produce more accurate predictions across a broader range of user-item combinations, thereby improving performance in the counterfactual test.
Comprehensive experiments validate the effectiveness of \proposed in both factual and counterfactual tests. 
Our implementation is accessible via: \url{https://github.com/SeongKu-Kang/BPL}.



\end{abstract}


\IEEEpeerreviewmaketitle

\section{Introduction}
\input{Sections/010Intro}

\section{Related Work}
\input{Sections/050Related}

\section{Concept and Problem Definition}
\label{Sec:concept}
\input{Sections/020Prelim_analysis}

\section{Preliminaries}
\input{Sections/021Analysis}

\section{Methodology}
\label{Sec:method}
\input{Sections/031Method}

\section{Experiments}
\input{Sections/040Experiment}

\section{Conclusion}
\input{Sections/060Conclusion}

\ifCLASSOPTIONcaptionsoff
  \newpage
\fi

\bibliographystyle{IEEEtran}
\bibliography{reference}

\end{document}

%% file: Sections/010Intro.tex
Real-world recommender systems form a feedback loop in which the systems' recommendations influence user behaviors, which in turn serve as training data for the system \cite{bias_survey}. 
This feedback loop leads to the creation and amplification of various biases affected by multiple factors, including but not limited to user selection patterns, item exposure mechanism, and influence of public opinions \cite{zhang2021tripartite, liu2016you}.
These biases progressively cause the training data to deviate from users' true preference, ultimately degrading the user satisfaction.

Over the past decade, debiasing learning has been actively studied for various user behaviors, such as explicit feedback~\cite{AT, STABLEDR, li2022tdr}, implicit feedback \cite{liu2024debiased, zhu2024mitigating}, and post-click conversion rate \cite{xu2022ukd}.
We focus on explicit feedback, which has additional challenges arising from its limited availability and the fine-grained levels of prediction targets (e.g., 1 to 5 rating scale).
Two foundational approaches are inverse propensity score (IPS) \cite{IPS} and data imputation \cite{AT}. 
IPS reweights data with propensity scores to correct the skewness of training data, while data imputation assigns missing ratings (or errors) for unrated data. 
Doubly robust (DR) \cite{DR} combines the two approaches to leverage their strengths with enhanced robustness, and recently many subsequent methods \cite{STABLEDR,li2022tdr,li2022multiple, song2023cdr, DCETDR} have further improved its effectiveness.

However, prior studies on debiasing learning have predominantly focused on the specialized (called \textit{counterfactual}) test environment, without considering its impact in the typical (called \textit{factual}) test environment. 
Specifically, test environments for recommender systems can be categorized as:
\begin{itemize}
    \item Factual test \cite{NCF, GLOCALK, berg2017graph} collects test feedback from \textit{actual interactions} between users and items. 
    It is the most widely used setup to assess a model's capability in predicting subsequent user behaviors.
    \item Counterfactual test \cite{IPS, AT, STABLEDR} collects test feedback via \textit{randomized controlled trials} (RCT), where system-induced biases are eliminated by exposing items randomly. 
    It is a specialized setup to assess a model's debiasing capability.
\end{itemize}
It is desirable to have a model that performs well in both tests rather than excelling in only one, as each test environment highlights the benefit of a specific aspect \cite{InterD};
the factual test emphasizes accurate predictions of subsequent behaviors, which are directly connected to service revenue, while the counterfactual test focuses on unbiased predictions, which influence user satisfaction in the long term.
A model with high accuracy in both tests can provide accurate predictions not only for randomly selected items (simulated by the counterfactual test) but also for those influenced by recommendation policies and popularity trends (simulated by the factual test), ultimately offering greater benefits for platform deployment.


\begin{figure*}[t]
\centering
\begin{subfigure}[t]{0.263\linewidth}
    \includegraphics[width=\linewidth]{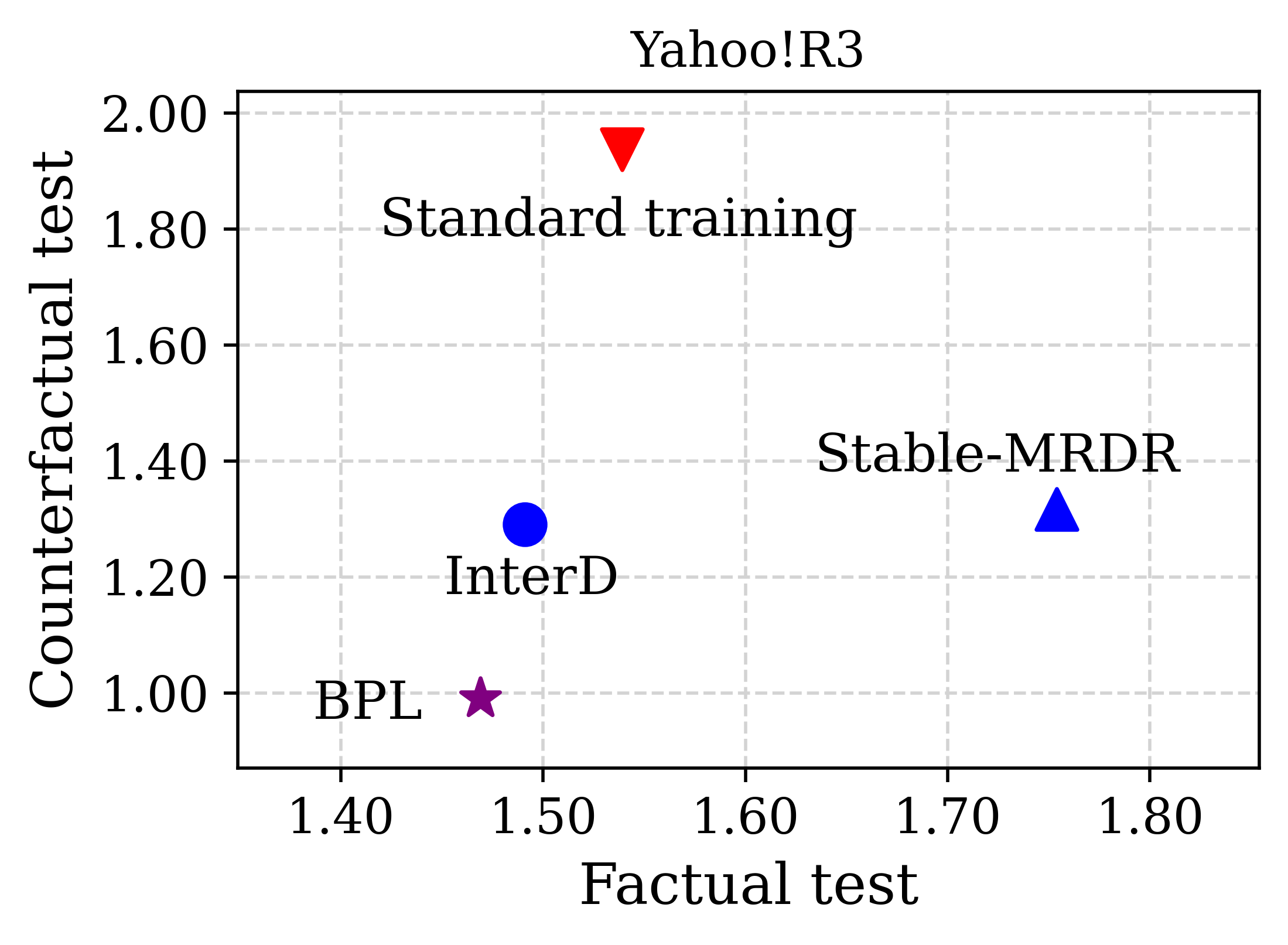}
\end{subfigure}
\hspace{0.3cm}
\begin{subfigure}[t]{0.263\linewidth}
    \includegraphics[width=\linewidth]{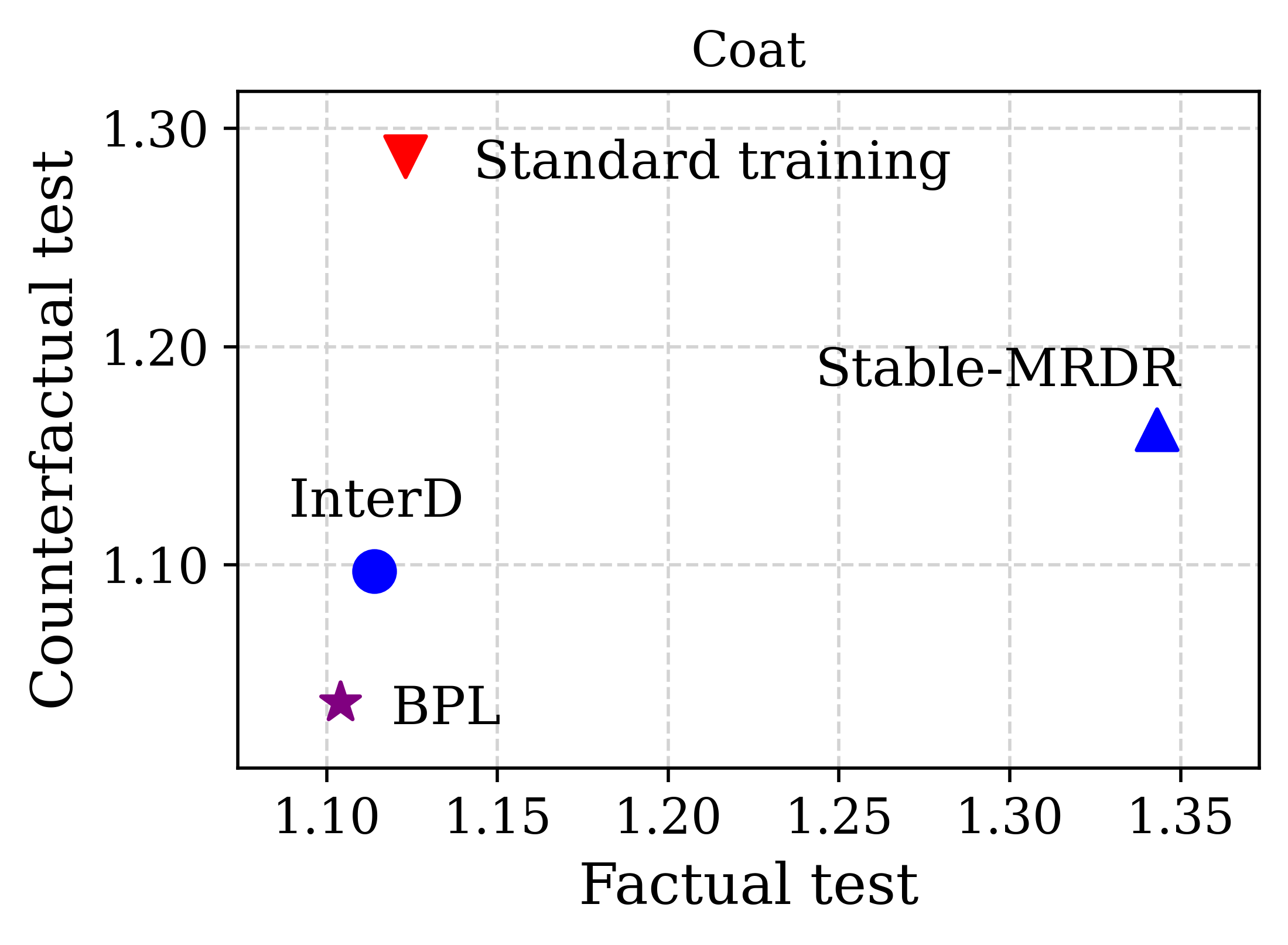}
\end{subfigure} 
\hspace{0.3cm}
\begin{subfigure}[t]{0.263\linewidth}
    \includegraphics[width=\linewidth]{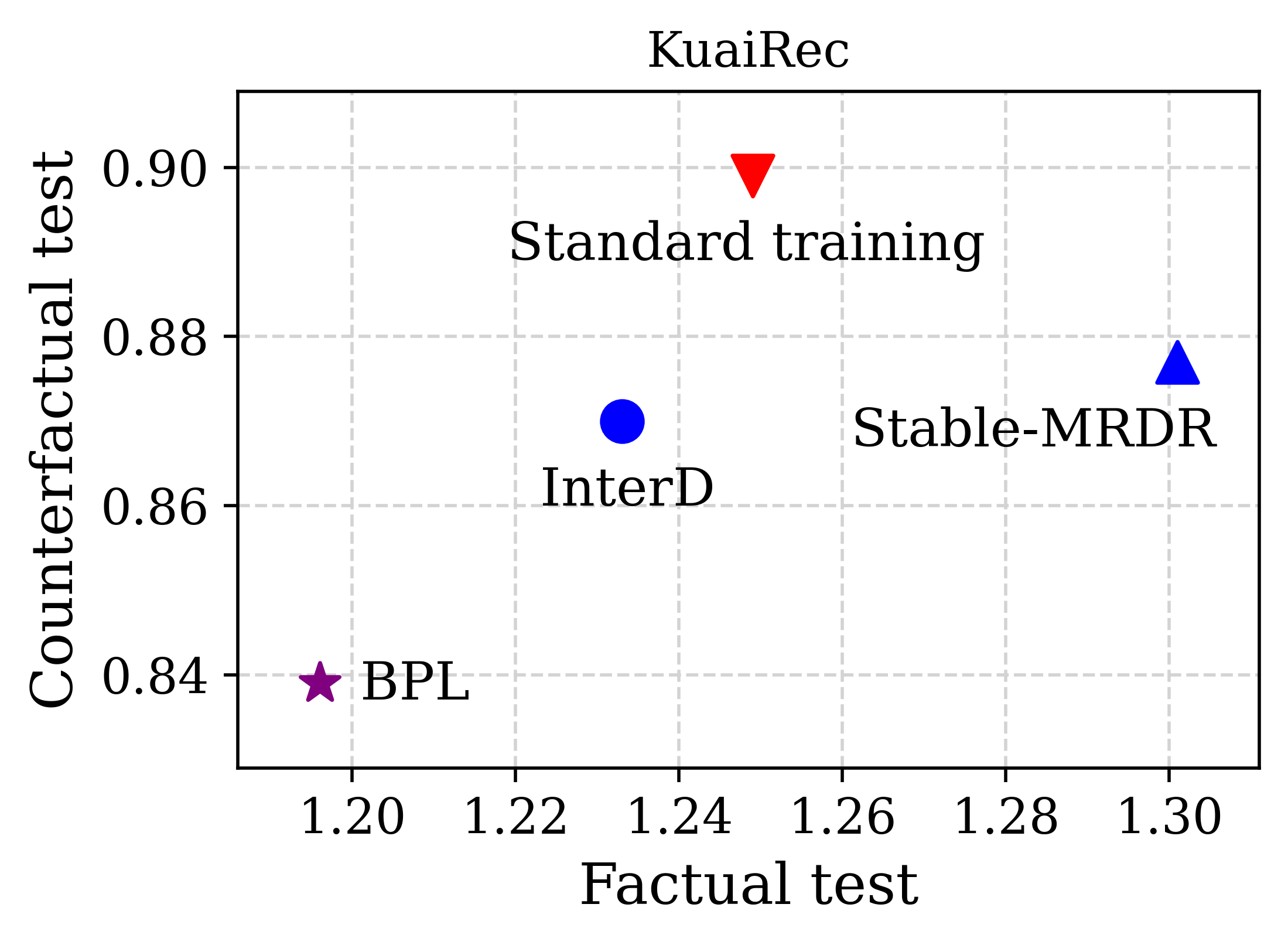}
\end{subfigure} 
\caption{Mean squared errors of various preference learning methods on factual and counterfactual test across three real-world datasets. 
We compare four different methods: (a) Standard training without debiasing learning, (b) Stable-MRDR \cite{STABLEDR}, a recent debiasing method, (c) InterD \cite{InterD}, a representative knowledge distillation method, and (Purple) \proposed, the proposed method. 
}
\label{fig:intro}
\vspace{-0.3cm}
\end{figure*}

Though effective in the counterfactual test, existing debiasing learning methods often significantly degrades accuracy in the factual test, resulting in a sub-optimal trade-off between the two test environments.
Figure~\ref{fig:intro} shows the performance of various methods in these environments.
Standard training produces a biased model that performs well in the factual test but poorly in the counterfactual test.
Stable-MRDR \cite{STABLEDR}, a representative debiasing learning method, improves performance in the counterfactual test but significantly degrades performance in the factual test.
Pointing out these distinct tendencies, a recent method, InterD \cite{InterD}, employs knowledge distillation (KD) from the biased and debiased models by interpolating their predictions.
While this approach improves the balance between the two test environments, its effectiveness remains limited by the quality of the pre-trained biased and debiased models used as teachers for distillation.
Consequently, this often results in only marginal improvements or, in some cases, performance degradation compared to the teacher models.
Furthermore, it is important to note that the teacher models themselves still have considerable room for improvement, as each performs well in only one of the two test environments.

In this work, we introduce a new preference learning framework, called \textbf{B}ias-adaptive \textbf{P}reference distillation \textbf{L}earning (\proposed), to gradually uncover underlying preference with dual distillation strategies.
First, we formulate our task as a risk minimization problem over all potential user-item combinations, based on risk minimization theory \cite{ben2010theory}.
We then decompose the risk upper bound into three distinct terms, each corresponding to: empirical risk on collected feedback, divergence between collected feedback and the remaining data, and the discriminability of ratings from learned representations. 
\proposed is composed of three major learning objectives, each tailored to minimize one of the upper-bound terms.

A core component of \proposed is its dual distillation strategies, designed to enhance rating-discriminability and minimize the third term, which has not been extensively studied in the previous literature.
Specifically, \proposed employs two types of distillation: 
(1) \textbf{reliability-filtered self-distillation}, which iteratively refines the target model to uncover preferences for unrated data. 
Utilizing the self-distillation approach \cite{allen-zhu2023towards}, we refine the model's knowledge based on its own predictions throughout the training.
Equipped with a filtering scheme to identify reliable predictions, the model progressively produces more accurate predictions for an expanding portion of the unrated data.
(2) \textbf{confidence-penalized preference distillation}, which supplements limited collected feedback using knowledge from the biased teacher model.
Employing a specialized form of teacher-student distillation~\cite{KD}, we seek to distill predictions that can be accurately inferred by the biased teacher.
To discouraging memorizing patterns that cannot be generalized to the remaining data, we combine the distillation loss with a confidence penalty, allowing the model to learn the teacher predictions while mitigating overconfidence.
These two distillation strategies are adaptively balanced based on the affinity of each potential user-item combination to the collected feedback, reflecting the predictive~capability~of~the~biased~teacher.

These dual distillation strategies enrich preference knowledge within representations, improving rating-discriminability and driving high performance in both factual and counterfactual test environments.
Through adaptive distillation from the biased teacher, \proposed retains accurate preference knowledge aligned with the collected feedback, leading to high performance in the factual test.
Also, rather than relying solely on pre-trained teachers, \proposed iteratively refines its knowledge through self-distillation.
This allows for progressively uncovering preferences across a broader range of user-item combinations, improving performance in the counterfactual test.
Our core contributions~are~as~follows:
\begin{itemize}[leftmargin=*]
    \item We introduce a new preference learning framework based on risk minimization theory, concurrently optimizing for both factual and counterfactual test environments.

    \item We propose dual distillation strategies: reliability-filtered self-distillation and confidence-penalized preference distillation, which are adaptively balanced based on the affinity to the collected feedback. 
    
    \item We validate the effectiveness of \proposed through extensive experiments.
    Also, we provide comprehensive analyses to evaluate the efficacy of each component.
\end{itemize}

%% file: Sections/050Related.tex
We introduce previous methods of debiasing learning for explicit feedback and KD, which are closely related to \proposed.\footnote{There are separate research lines for debiasing other behavior types (e.g., implicit feedback, click-through rate), and we guide readers to \cite{bias_survey}.}

\smallsection{{Debiasing learning for explicit feedback.}}
Debiasing learning has been actively studied for recommendation.
IPS \cite{IPS, AT, DRLTD} reweights each data with propensity scores to correct the skewness of the training set.
The propensity can be computed by using statistics \cite{AT} or dedicated models \cite{DRLTD, li2022multiple}.
Also, data imputation \cite{AT}, which assigns pseudo-labels to unrated data, has been widely studied to learn beyond the biased training set.
Furthermore, there have been attempts to utilize special training set \cite{autodebias, DRLTD, li2023balancing, li2023propensity, agarwal2019general}, collected by random exposure of items. 
With the special training set, prior methods have improved the quality of propensity estimation \cite{DRLTD}, imputation value \cite{DRLTD, autodebias}, and also achieve a broad debiasing effect~\cite{autodebias}.
However, obtaining such data can be challenging in real-world applications, as the random exposure policy inevitably degrades user satisfaction~\cite{kweon2024doubly, saito2022towards}.

On the one hand, several methods \cite{saito2022towards, chen2020esam, pan2023discriminative} have formulated the debiasing task as a domain adaptation problem handling distribution shift between the rated (or popular) data and the unrated (or unpopular) data.
Employing an adversarial learning approach, they have effectively reduced the divergence between two distributions, mitigating the impact of biases.
Furthermore, \cite{pan2023discriminative} proposes a new label-conditional clustering to improve representation discriminability.
There have also been attempts to directly model the rating process of exposure, selection, and evaluation steps \cite{zhang2021tripartite}, and to use invariant learning to capture environment-invariant preference \cite{zhang2023invariant}.
A notably dominant approach is DR \cite{DR}, which combines IPS and data imputation approaches.
\textcolor{red}{Many subsequent methods \cite{STABLEDR, song2023cdr, zhang2024uncovering, li2024debiased, kweon2024doubly, li2025meta} have further improved its effectiveness and achieved remarkable performance in the counterfactual test environment.}
However, they often largely degrade accuracy in the factual test environment.
\proposed does not rely on unbiased training data, and is designed to achieve low errors in both factual and counterfactual tests.

\smallsection{{Knowledge distillation.}} 
KD is a technique to transfer knowledge from a pre-trained teacher model to improve a target model \cite{KD}.
For recommender system, KD has been extensively studied to generate a lightweight model that preserves the performance of a massive teacher model \cite{hetcomp, chen2023unbiased, DERRD}.
Leveraging the teacher as ground-truth knowledge sources, existing methods train the target model to mimic the final predictions \cite{chen2023unbiased, hetcomp} and intermediate representations \cite{DERRD} from the teacher.
On the other hand, a few recent methods \cite{liu2022kdcrec, liu2020general} utilize distillation with unbiased training data for debiasing.
However, as mentioned earlier, a biased model and a debiased model show a clear trade-off in factual and counterfactual test environments, and relying on each of them inevitably leads to poor performance in one of the test environments.
Pointing out this problem, InterD \cite{InterD} introduces a new distillation method that combines the knowledge of biased and debiased models, and transfers it to the target model.
While it achieves a better trade-off compared to using one of the models, its effectiveness is often bounded by its teacher models; according to our experiments in Section \ref{subsec:dual}, it often fails to bring large improvements compared to the debiased teacher in counterfactual test. 

%% file: Sections/020Prelim_analysis.tex
\smallsection{Definition 1} (Recommendation with explicit feedback).
Let $\mathcal{U}$ and $\mathcal{I}$ denote the set of users and items, respectively.
We focus on explicit feedback where a user $u$ expresses their preference for an item $i$ through multi-level ratings $r_{ui} \in \mathcal{R}$, with $\mathcal{R} = \{1, \ldots, K\}$ and higher values indicating stronger preferences.
$\mathcal{S} = \mathcal{U} \times \mathcal{I}$ denotes the space of all user-item pairs\footnote{In this work, we use the terms ``data'' and ``pairs'' interchangeably.}, which can be divided into a subspace of rated pairs $\mathcal{S}^{1}$ and a subspace of unrated pairs $\mathcal{S}^{0}$.
Let $s_{ui}$ denote a Bernoulli random variable representing the presence of $(u,i)$ pair in $\mathcal{S}^1$, i.e., $s_{ui}=1$ if $(u,i) \in \mathcal{S}^1$, otherwise $0$.
Our task is to predict the potential outcome when $s_{ui}$ had been set to $1$, i.e., if an item $i$ had been rated by a user $u$, what would be the feedback?

\smallsection{Definition 2} (Preference learning as risk minimization).
Let $f:\mathcal{U}\times\mathcal{I}\rightarrow \mathcal{R}$ denote a recommendation model.
If ratings are provided for all user-item pairs in \s, we can train the model to minimize the following \textit{true risk}: 
\begin{equation}
    \mathcal{L}_{true} = |\mathcal{S}|^{-1}\sum_{(u,i)\in \mathcal{S}} \ell(f(u,i), r_{ui}),
\end{equation}
where $\ell(\cdot,\cdot)$ denotes the error function. 
\textcolor{red}{We refer to knowledge that can minimize this true risk, which allows for accurate recommendations for all user-item pairs, as \textit{true preference}.}

As this true risk is not accessible, the model is typically optimized to minimize the empirical risk: $\mathcal{L}_{emp} = |\mathcal{S}^1|^{-1}\sum_{(u,i)\in \mathcal{S}^1} \ell(f(u,i), r_{ui})$.
\textcolor{red}{However, due to various biases, this approach falls short of capturing the true preference.
To elaborate, the collected ratings are missing-not-at-random as users do not rate items uniformly; 
users tend to select items that they like and also actively rate particularly bad or good items (selection bias) \cite{hernandez2014probabilistic}. 
The rating process is also affected by various factors such as item popularity and recommendation policy (exposure, position bias) \cite{expoMF, agarwal2019general}.
Additionally, user judgment is influenced by public opinions, leading to different distributions when users rate items before or after being exposed to public opinions~(conformity~bias)~\cite{liu2016you}.
}




\smallsection{Definition 3} (Factual and counterfactual test).
Test environments for recommender systems can be grouped into two categories \cite{InterD}:
(1) Factual test collects feedback from actual user-item interactions.
(2) Counterfactual test collects feedback via randomized controlled trials (RCT), where system-induced biases are eliminated by exposing items randomly.
\textcolor{red}{Because the factual test relies on actual interactions, popular items that users tend to engage with more frequently are often overrepresented in the evaluation.
In contrast, the counterfactual test offers an alternative evaluation perspective to a broader range of items, including niche or less popular ones.}


Note that the user-item pairs collected in each test complementarily cover $\mathcal{S}$:
the factual test covers unseen pairs closer to the collected feedback (i.e., \so), while the counterfactual test includes unseen pairs more widely distributed across the space.
A model that achieves high accuracy in both tests can provide accurate recommendations for items selected both randomly and through various factors (e.g., popularity trends), offering greater benefits to the platform.


\smallsection{Problem formulation}.
Given a set of rated data in \so, our goal is to generate a model that minimizes the true risk, performing well on both factual and counterfactual test environments.

%% file: Sections/021Analysis.tex
\subsection{\textbf{Risk upper bound on} \s}
\label{subsec:philosophy}
To develop a systematic approach to minimize true risk, we first define the upper bound of the expected errors using risk minimization theory \cite{ben2010theory}. 
As this theory is based on properties of learned representations, we view the recommendation model as the combination of an encoder and a predictor $f=\eta \circ \phi$, where the encoder $\phi: \mathcal{U}\times \mathcal{I} \rightarrow \mathcal{Z}$ encodes each pair into the representation space, and the predictor $\eta: \mathcal{Z} \rightarrow \mathcal{R}$ predicts ratings from the representation.

\smallsection{Theorem 1 \cite{ben2010theory}} 
\textit{Let $\mathcal{H}$ be the hypothesis class of predictors on representation space. 
For $\eta \in \mathcal{H}$, let $\epsilon_{\mathcal{S}}(\eta)$ denote the expected errors on \s, i.e., $\epsilon_{\mathcal{S}}(\eta) = |\mathcal{S}|^{-1}\sum_{(u,i)\in \mathcal{S}}\ell(f(u,i), r_{ui})$.
$\mathcal{Z}_{\mathcal{S}^0}$ and $\mathcal{Z}_{\mathcal{S}^1}$ denote the distributions of \sz and \so over $\mathcal{Z}$, respectively.
Given $\lambda_0=\frac{\mid \mathcal{S}^0 \mid}{\mid \mathcal{S}^{\,\,\,}\mid}$, which is the ratio of unrated data in $\mathcal{S}$, $\forall \eta \in \mathcal{H}$:}
\begin{equation}
\label{eq:risk}
\begin{aligned}
\epsilon_{\mathcal{S}}&(\eta) \leq 
 \epsilon_{\mathcal{S}^1}(\eta) \\ &+ \lambda_0 \left(\frac{1}{2}d_{\small{\mathcal{H}\Delta\mathcal{H}}}(\small{\mathcal{Z}_{\mathcal{S}^0}}, \small{\mathcal{Z}_{\mathcal{S}^1})} + (\epsilon_{\mathcal{S}^0}(\eta^*) + \epsilon_{\mathcal{S}^1}(\eta^*)) \right)
\end{aligned}
\end{equation}
The upper bound of the expected error consists of three terms:
\begin{enumerate}[leftmargin=*]\vspace{-\topsep}
    \item [\textbf{T1}:]$\epsilon_{\mathcal{S}^1}(\eta)$: the error on rated space \so.
    \item [\textbf{T2}:] $d_{\mathcal{H}\Delta\mathcal{H}}(\mathcal{Z}_{\mathcal{S}^0}, \mathcal{Z}_{\mathcal{S}^1})$: the $\mathcal{H}\Delta\mathcal{H}$-divergence \cite{ben2010theory} which is defined~as $2 \sup _{\eta, \eta^{\prime} \in \mathcal{H}}\mid p_{z \sim \mathcal{Z}_{\mathcal{S}^0}}\left[\eta(z) \neq \eta^{\prime}(z)\right]-p_{z \sim \mathcal{Z}_{\mathcal{S}^1}}\left[\eta(z) \neq \eta^{\prime}(z)\right] \mid$.
    It indicates the discrepancy between \sz and \so over $\mathcal{Z}$. 
    \item [\textbf{T3}:] $\epsilon_{\mathcal{S}^0}(\eta^*) + \epsilon_{\mathcal{S}^1}(\eta^*)$: the combined error of the ideal hypothesis, i.e., $\eta^* = \argmin_{\eta\in \mathcal{H}}(\epsilon_{\mathcal{S}^0}(\eta) + \epsilon_{\mathcal{S}^1}(\eta))$, on~\sz~and~\so.
\end{enumerate}
\noindent
T1 can be minimized by standard training with ground-truth ratings, and T2 can be approximated and minimized by adversarial learning \cite{ ganin2016domain}.
Combining standard training (T1) and adversarial learning (T2) has achieved huge success in handling unlabeled data for domain adaptation \cite{ganin2016domain} and has also been applied for recommender systems \cite{saito2022towards, chen2020esam, pan2023discriminative}.

The remaining term is T3, which is the main focus of this paper.
T3 measures discriminability of representations \cite{chen2019transferability, pan2023discriminative}, which indicates the easiness of distinguishing different ratings (i.e., $\mathcal{R}$) by a supervised predictor trained over the representations.
To reduce T3, the representations should contain sufficient preference information that allows for distinguishing ratings based on them. 
Unlike T1 and T2, directly approximating T3 is infeasible due to the absence~of~ratings~in~\sz.



\subsection{\textbf{Motivational Analyses}}
\label{subsec:anal}
We present our analysis showing that the existing learning strategies have limitations in minimizing risk across \s.
Furthermore, we discuss how a biased model can aid in learning more accurate preference.
We train the recommendation model using the biased training set and report the results on the RCT test set. 
The results on Yahoo!R3 dataset are reported, with similar findings observed across other datasets.

\smallsection{\textbf{\so-affinity}.}
\label{subsec:s1affinity}
As users continue to utilize the system, some data in \sz will be naturally rated.
Given that ratings are missing-not-at-random, unrated data have varying probabilities of being rated potentially, i.e., $p(s_{ui}=1)$.
This probability reveals how similar each data point is to the already collected feedback, and we leverage it for analysis.
To estimate the probability, we train a binary classifier to distinguish whether each pair $(u,i)$ comes from \sz or \so using the binary cross-entropy loss.\footnote{$\mathcal{L} = -\sum_{(u,i)\in \mathcal{S}}s_{ui} \log \hat{p}(s_{ui}=1) + (1-s_{ui})\log (1-\hat{p}(s_{ui}=1))$. We use a one-layer MLP with a sigmoid activation function \textcolor{red}{(Appendix A)}.}
We refer to $\hat{p}(s_{ui}=1)$ as \so-affinity of each unrated data. 

\smallsection{\textbf{Analysis 1: Impact of learning strategy on prediction errors}.}
\label{subsubsec:errors}
We first analyze the model prediction errors from four different learning strategies: \textbf{(1)} the standard training on \so (T1), \textbf{(2)} the standard training with adversarial learning (T1 + T2), \textbf{(3)} the representative debiasing learning with doubly-robust (DR) \cite{STABLEDR}, \textbf{(4)} \proposed, the proposed approach, designed to minimize all three terms of the risk upper bound.
Note that DR employs different approaches that leverage propensity scores, and is not directly linked with the three terms in Eq. \ref{eq:risk}.

\begin{figure}[t]
\centering
\begin{subfigure}[t]{\linewidth}
\centering
    \includegraphics[height=3.8cm]{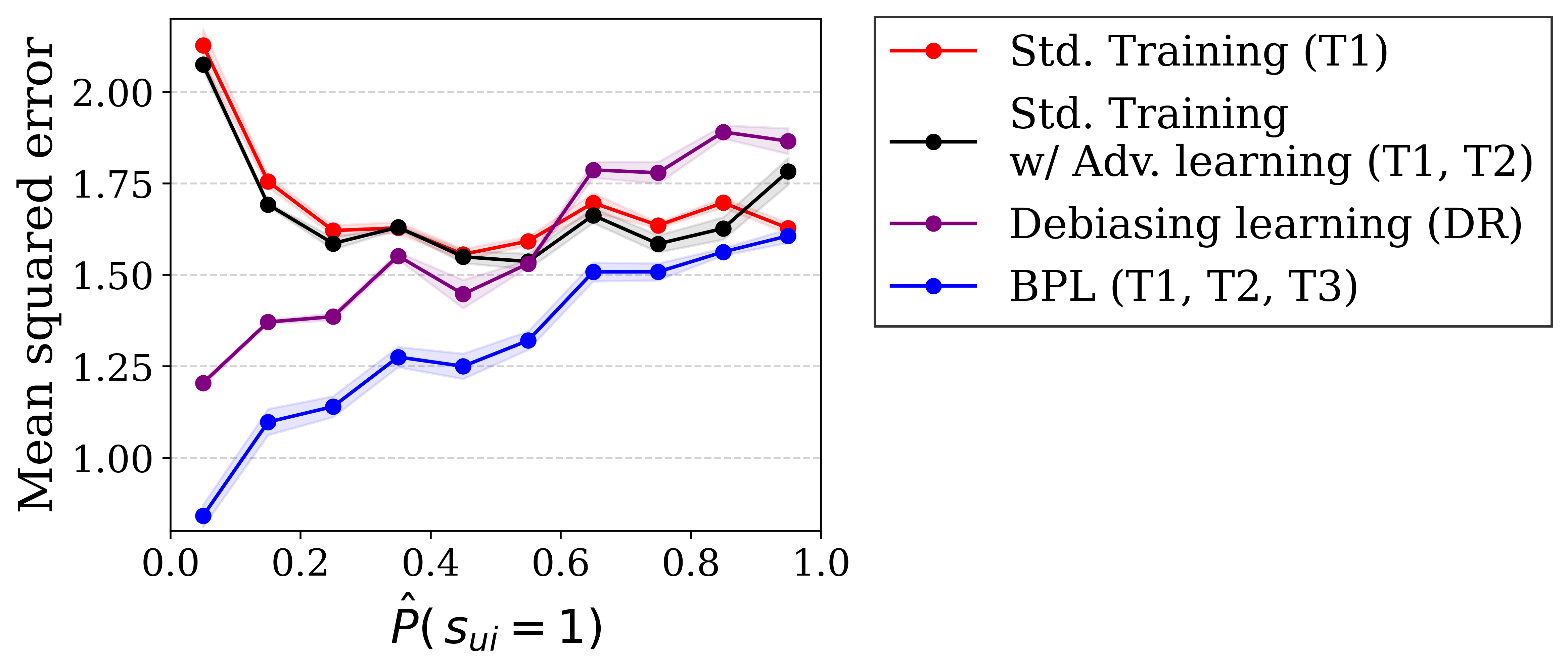}
\end{subfigure}
\caption{Relationships between \so-affinity (i.e., $\hat{p}(s_{ui}=1)$) and model prediction errors with varying learning strategies.} 
\label{fig:motivate_mse}
\vspace{-0.3cm}
\end{figure}

Figure \ref{fig:motivate_mse} presents prediction errors according to the affinity.
We observe that the debiasing learning and standard training show highly different tendencies.
For data with low affinity, debiasing learning significantly reduces the errors.
However, this improvement comes at the cost of largely degraded accuracy on high-affinity data.
DR employs propensity scores that reweight each data to prevent over-reliance on the collected feedback, which can rather hinders learning on high-affinity data.
On the other hand, while the standard training generates a biased model with high errors for data with low affinity, it consistently achieves low errors for a substantial amount of high-affinity data.
Lastly, although adversarial learning reduces errors, its effectiveness remains limited, indicating that simply minimizing overall divergence is insufficient.


These observations reveal two insights: 
(i) The biased model’s knowledge of high-affinity data holds potential for improvement by complementing the limited feedback and enriching the supervision for preference learning.
(ii) The \so-affinity score provides a strategic foundation for the selective utilization of biased model knowledge by distinguishing the data that the biased model can aptly comprehend.

\begin{figure}[t]
\centering
\hspace{-0.7cm}
\begin{subfigure}[t]{0.5\linewidth} 
    \includegraphics[height=3.8cm]{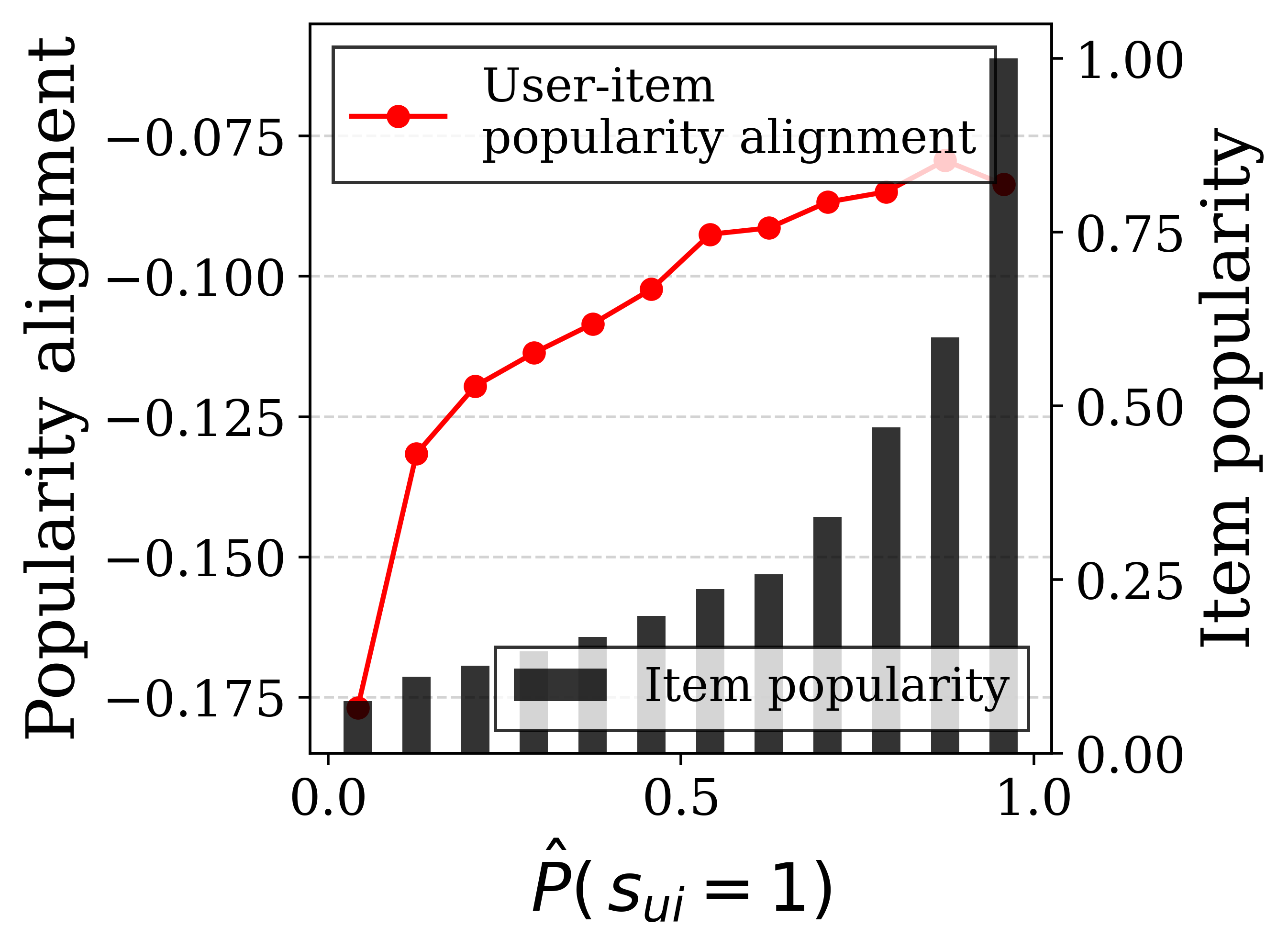}
\end{subfigure}
\hspace{0.5cm}
\begin{subfigure}[t]{0.3\linewidth}
    \includegraphics[height=3.8cm]{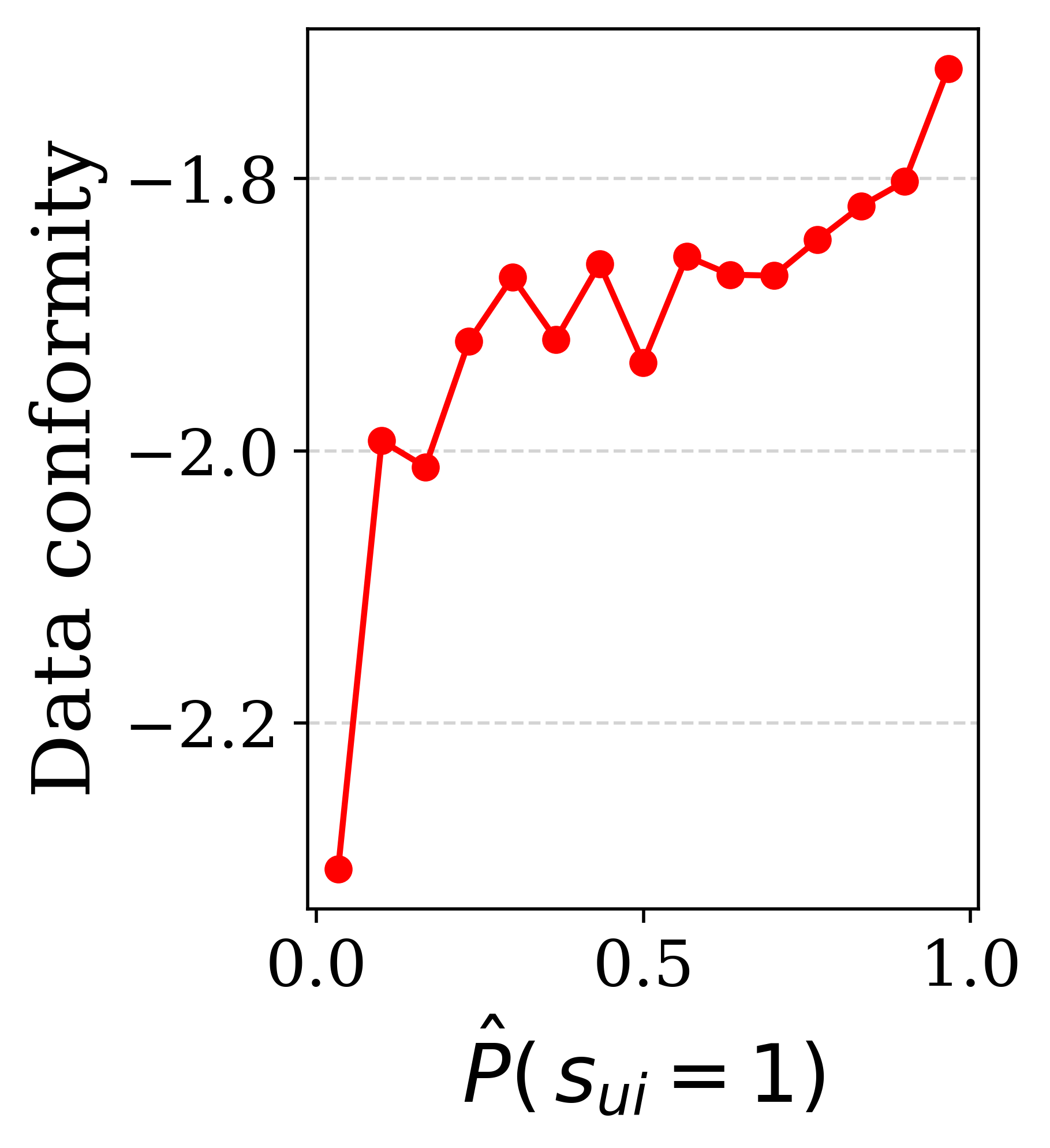}
\end{subfigure}
\caption{Relationships between \so-affinity (i.e., $\hat{p}(s_{ui}=1)$) and three factors that affect rating process.} 
\label{fig:motivate_factors}
\vspace{-0.4cm}
\end{figure}





\smallsection{\textbf{Analysis 2: $\mathcal{S}^1$-affinity and rating process-related factors}.}
\label{subsec:anal_setup}
To gain a deeper insight into why the biased model achieves low errors on high-affinity data, we explore the relationships between \so-affinity and three factors that can affect users' rating process\footnote{It is infeasible to specify individual factors that affect the rating process. We analyze three factors that are widely recognized in prior literature.}:
\textbf{(1)} \textbf{item popularity} denotes the normalized interaction frequency of each item. It is well known that popularity strongly influences the exposure and selection process \cite{zhang2021tripartite}.
\textbf{(2)} \textbf{user-item popularity alignment} denotes the alignment between the average popularity of the user's historical items and the popularity of each item.\footnote{$\operatorname{sim}(pop(u), pop(i))$, where $pop(u)=\operatorname{AVERAGE}(\{pop(i)\}_{s_{ui}=1})$. 
We use the negative squared difference for the similarity function, i.e., $\operatorname{sim}(x,y) = -(x-y)^2$. }
This factor jointly considers both user-side and item-side popularity.
For instance, ``blockbuster fan-indie film'' has a low alignment value due to their dissimilar popularity levels.
\textbf{(3)} \textbf{data conformity} measures how closely each user's opinion aligns with the public opinion \cite{liu2016you}.\footnote{$\operatorname{sim}(r_{ui},\Bar{r}_i)$, where $\Bar{r}_i$ is the averaged training rating for item $i$.}
For all three factors, a higher value indicates greater popularity, alignment, and conformity.

Figure~\ref{fig:motivate_factors} presents the results.
Both item popularity and popularity alignment show strong positive correlations with \so-affinity.
This shows that both \so and high-affinity data lean towards popular items and users who prefer popular items, and they are likely to be similarly affected by popularity-related factors.
Also, data conformity shows a strong correlation with \so-affinity, suggesting that public opinion generally has a stronger impact on the evaluation of high-affinity data, compared to data deviating from \so.
While specifying the impacts of each bias is not our focus, a possible explanation is that users tend to encounter public opinion more frequently for popular items. 
These results show that \textit{\so and high-affinity data have similar characteristics in terms of various factors affecting the users' rating process}, which also supports the low errors on high-affinity data by the biased model.

These analyses motivate us to selectively leverage the biased model’s knowledge, which aids in predicting user preferences for high-affinity data, thereby enhancing preference learning.


%% file: Sections/031Method.tex
\begin{figure*}[t]
    \centering
    \includegraphics[width=0.95\textwidth]{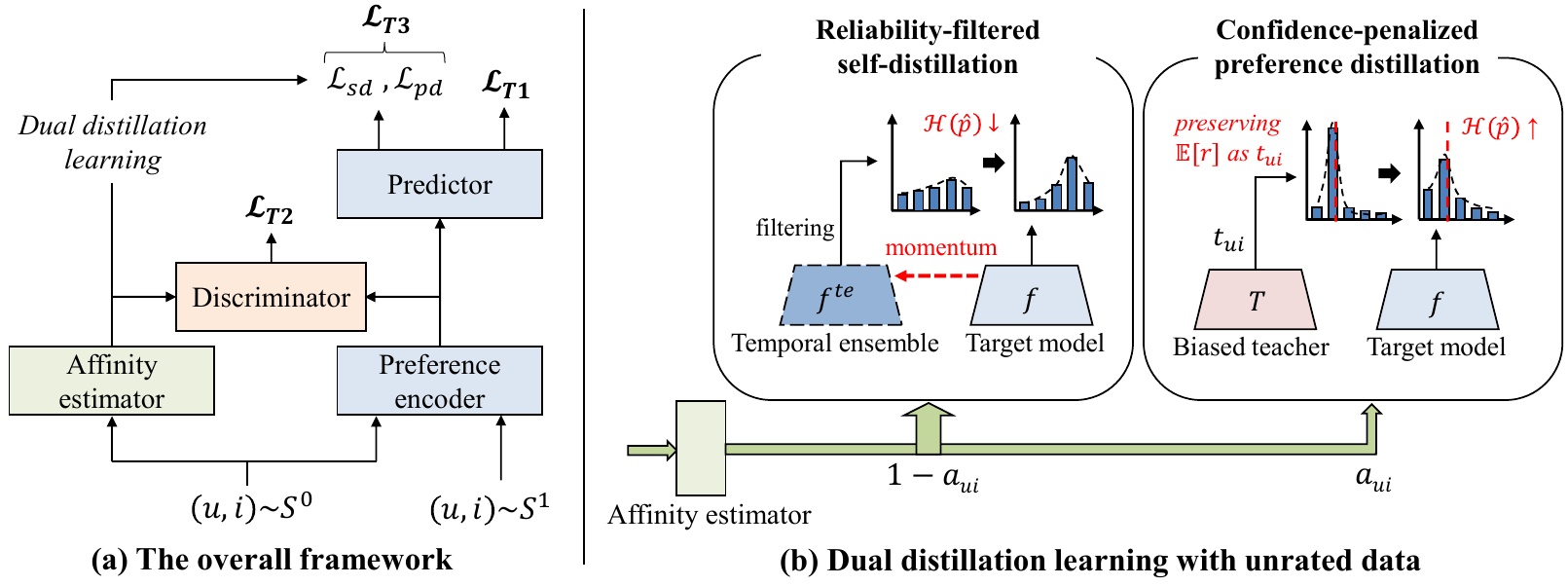}
    \caption{Illustration of \proposed. (a) The overall framework, (b) Dual distillation learning with unrated data.
    Best viewed in color.}
    \label{fig:method}
    \vspace{-0.3cm}
\end{figure*}

We propose \proposed, designed to effectively reduce the risk upper bound on user-item space.
\proposed employs three major learning objectives, each of which focuses on reducing each term of the upper bound. 
Specifically, \proposed trains the model using $\mathcal{L}_{T1}$ to learn preference from rated data (Section\ref{subsec:methodt1}), $\mathcal{L}_{T2}$ to reduce the divergence between \sz and \so in the representation space (Section \ref{subsec:methodt2}), and $\mathcal{L}_{T3}$ to discover preference for unrated data via dual distillation strategies that adaptively leverage the biased model knowledge (Section \ref{subsec:methodt3}).
These three objectives are jointly optimized to reduce the upper bound (Section \ref{subsec:method_train}).
Figure \ref{fig:method} presents the overview of \proposed.

\subsection{\textbf{Preference learning with rated data (T1)}}
\label{subsec:methodt1}
We first learn preference with ground-truth ratings from users, by minimizing T1.
For encoder $\phi$ of the model $f$, various architectures can be flexibly employed, from ID-based \cite{koren2009matrix} to graph-based \cite{berg2017graph} encoder.
For predictor $\eta$, we use a $K$-class softmax classifier, where each class corresponds to each rating level, and $\hat{p}(r_{ui}=k)$ denotes the softmax probability for class $k$.
We use the supervised~loss~as:
\begin{equation}
\begin{aligned}
\label{eq:Lp}
\min_{\phi, \eta}\mathcal{L}_{T1} =|\mathcal{S}^1|^{-1}\sum_{(u,i)\in \mathcal{S}^1} \ell_{s}(f(u,i), r_{ui}),
\end{aligned}
\end{equation}
where $\ell_s = - \sum_{k\in\{1,...,K\}} \mathbbm{1}[r_{ui}=k] \log \hat{p}(r_{ui}=k)$.
The predicted rating $\hat{r}_{ui}$ is obtained by the expected value based on the softmax probabilities \cite{berg2017graph}: 
$\hat{r}_{ui}=\mathbb{E}_{\hat{p}}[r_{ui}]=\sum_{k} k \,\hat{p}(r_{ui}=k)$.
This choice allows us to readily control the prediction confidence while preserving the accuracy, which will be explained in Section \ref{subsub:kd}.

\textcolor{red}{As discussed in Section~\ref{subsec:anal}, due to various biases in \so, this approach often falls short of capturing true preferences.
The remaining parts of \proposed are dedicated to alleviating this problem based on the risk upper bound (Section~\ref{subsec:philosophy}).}

\subsection{\textbf{Distribution alignment learning (T2)}}
\label{subsec:methodt2}
T2 corresponds to the divergence between \sz and \so in the representation space.
We adopt adversarial learning which has been extensively studied in domain adaptation \cite{ganin2016domain, pan2023discriminative}.
It trains a discriminator to distinguish representations from different domains, while training the encoder to generate representations that the discriminator cannot distinguish.
By doing so, the encoder reduces the divergence between two domain distributions in the representation space~\cite{ganin2016domain}.

However, we empirically found that treating \sz and \so as independent domains hinders effective optimization.
In fact, two spaces are not inherently independent; as users continue to utilize the system, some data in \sz will be rated.
Naturally, the discriminator struggles to find a clear decision boundary that separates high-affinity data from \so.
Such degraded discrimination capability subsequently limits the encoder's ability to achieve alignment, thereby hindering effective optimization.
This issue is consistent with the findings in \cite{jin2021re} that training a discriminator to distinguish already well-aligned representations leads to suboptimal results.

\textcolor{red}{As a solution, we borrow the idea from \cite{jin2021re}, which relabels a small set of samples near the decision boundary in adversarial learning.
Specially, we expand \so by adding a small subset of unrated data with high \so-affinity.}
Let \szo denote a set of unrated data with the highest $x\%$ affinities.
Given a representation $z_{ui}$, we train a discriminator $f_d: \mathcal{Z} \rightarrow [0,1]$ to distinguish whether $z_{ui}$ comes from the expanded space \so $\cup$ \szo or the remaining unrated data.
The encoder $\phi$ is trained to generate representations that prevent $f_d$ from~distinguishing~them:
\begin{equation}
\begin{aligned}
\label{eq:Ld}
    \min_{\phi}\max_{f_d} \mathcal{L}_{T2} = &\sum_{(u,i)\in \mathcal{S}^1 \cup \mathcal{S}^{01}}  \log (f_d(z_{ui}))  \\&+ \sum_{(u,i)\in \mathcal{S}^{0}\setminus \mathcal{S}^{01}}  \log (1-f_d(z_{ui}))
\end{aligned}
\end{equation}
$\mathcal{L}_{T2}$ corresponds to the empirical $\mathcal{H}$-divergence between two distributions, which empirically approximates $\mathcal{H}\Delta\mathcal{H}$-divergence \cite{ganin2016domain}.
We identify unrated data most likely to be rated potentially and treat them as (pseudo) rated data.
\textcolor{red}{This strategy enables the discriminator to find a more separable decision boundary, which in turn makes the encoder better align the distributions \cite{jin2021re}.}

\subsection{\textbf{Dual distillation learning with unrated data (T3)}}
\label{subsec:methodt3}
As discussed in \cref{subsec:philosophy}, T3 reflects the easiness of discerning ratings from representations.
That is, representations should include sufficient preference information that allows for distinguishing ratings based on them. 
However, due to the skewness and limited availability of ground-truth ratings, relying on $\mathcal{L}_{T1}$ falls short of finding preference for unrated data.

As a solution, we introduce dual distillation strategies that iteratively refine the target model, aided by the adaptive usage of biased model knowledge.
We first introduce \textbf{reliability-filtered self-distillation} (Section \ref{subsub:st}) to uncover preference for unrated data by iteratively refining the model based on its predictions. 
To facilitate accurate preference discovery, we propose \textbf{confidence-penalized preference distillation} (Section \ref{subsub:kd}) that supplements limited ground-truth ratings with the aid of a biased teacher model. 
The two distillations are balanced based on \so-affinity of each data (Section \ref{subsub:combine}).

\subsubsection{\textbf{Reliability-filtered self-distillation}}
\label{subsub:st}
Self-distillation is a special type of distillation where a model is used to generate ``soft labels'' for its own training data \cite{allen-zhu2023towards, wu2024teacher}.
Instead of using external teacher models, the model learns from its own predictions, typically probability distributions over classes, iteratively refining its knowledge.
This approach has proven effective in improving generalization by directly leveraging previously discovered hidden patterns that are not directly revealed by labeled data \cite{allen-zhu2023towards}.
Employing the self-distillation approach, we aim to generate predictions for unrated data and refine the model's preference knowledge over time.

However, unlike typical classification tasks, the rated data is highly limited and skewed, causing the model to generate many unreliable predictions.
To address this problem, we introduce a \textit{filtering scheme} to identify reliable predictions.
Through filtering, we selectively refine the model's preference knowledge by leveraging only the reliable predictions.
As training progresses, the model increasingly produces accurate predictions for a growing number of unrated data, discovering preferences across a broader unrated space.

\begin{figure}[t]
    \centering
    \includegraphics[height=3cm]{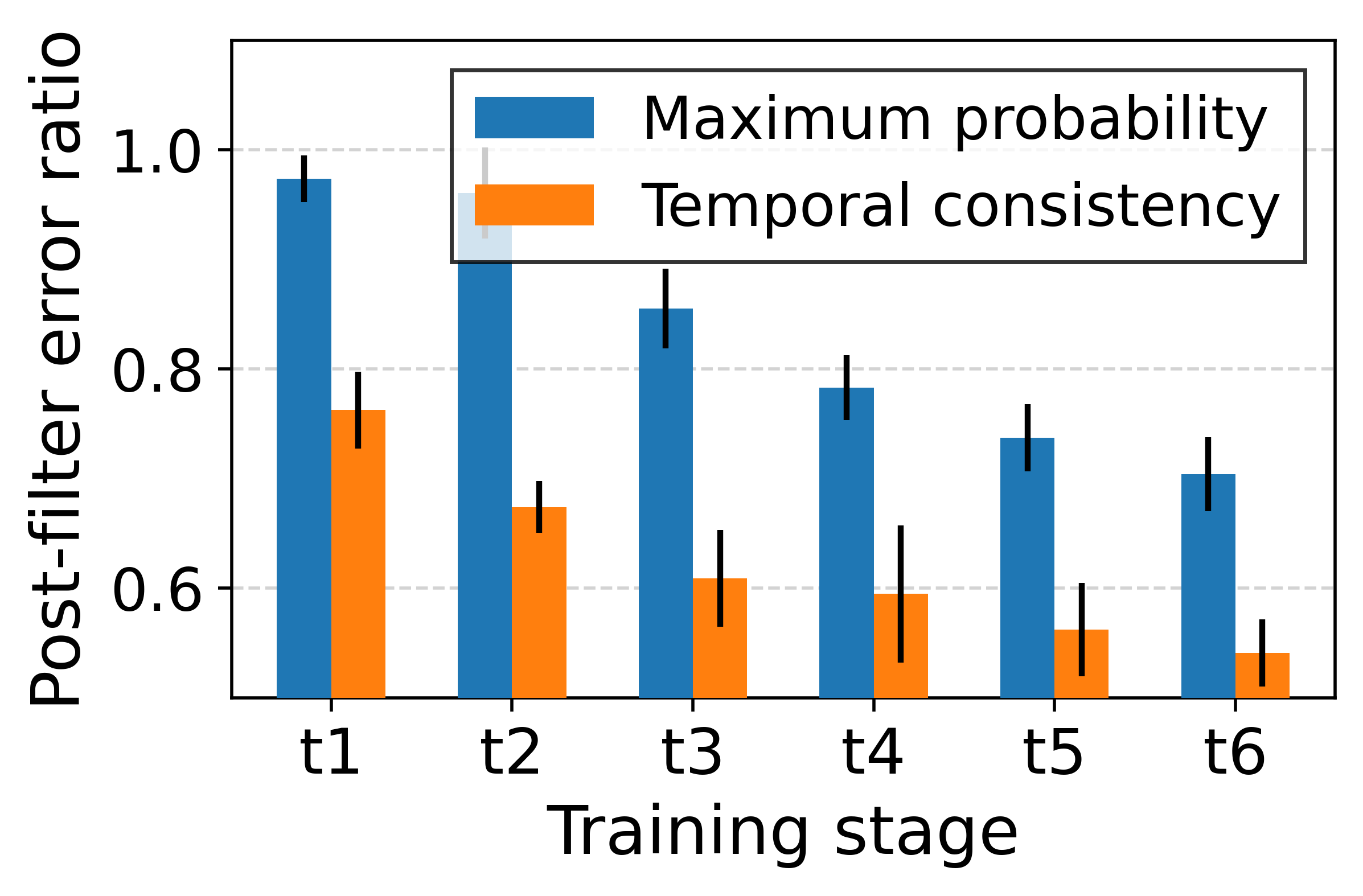}
    \hspace{0.1cm}
    \includegraphics[height=3cm]{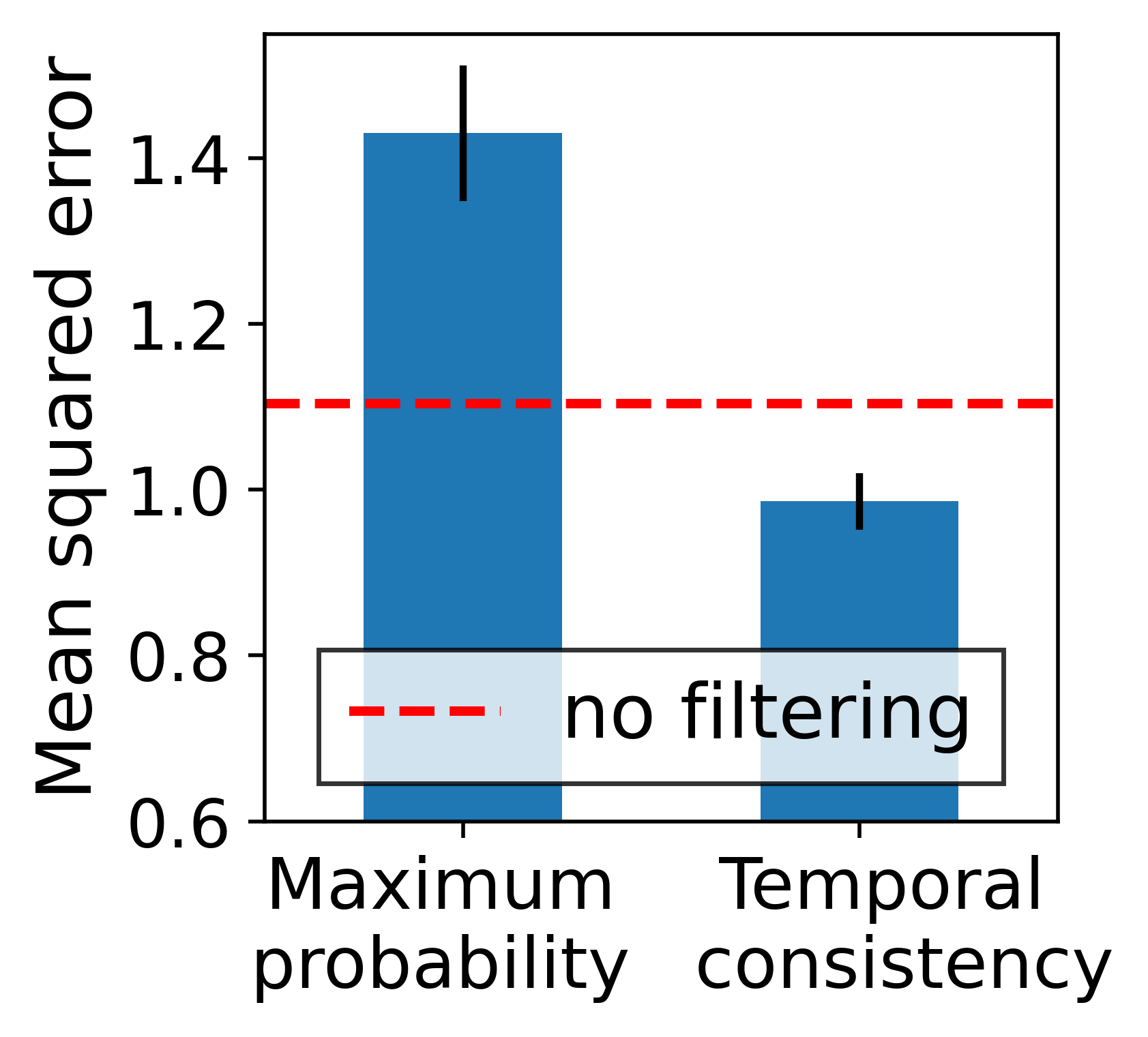}
    \caption{Comparison of filtering effects. (left) post-filtering error ratio, which denotes the ratio of average errors in the filtered dataset to the average errors in the total dataset, (right) recommendation performance. Results on Yahoo!R3-RCT.}
    \label{fig:filtering}
    \vspace{-0.3cm}
\end{figure}

\smallsection{Reliability filtering.}
As not all predictions are accurate, it is crucial to leverage reliable ones selectively.
To develop an effective filtering scheme tailored to \proposed, we examine two distinct metrics that have proven effective for assessing the reliability of model predictions:
\begin{itemize}[leftmargin=*]
    \item \textbf{Maximum probability} is a widely used metric \cite{freematch}, where a prediction is considered reliable if its maximum class probability exceeds a certain threshold, i.e., $\max_k \hat{p}(r_{ui}=k) > m$.
    We tune the threshold $m$ from $0.1$ to $0.9$ based on validation performance.
    \item \textbf{Temporal consistency} assesses how consistent the prediction is during training, supported by findings that inconsistent predictions that frequently change often contain large errors \cite{zhou2020time}.
    To evaluate consistency, we employ a temporal ensemble of the model, as done in \cite{zhou2020time}. 
    This temporal ensemble $f^{te}$ is updated as the moving average of $f$ with a discount factor $\tau \in [0,1]$, i.e., $\theta_{f^{te}} = \tau\theta_{f^{te}} + (1-\tau)\theta_f$.
    By setting a large value for $\tau$, $f^{te}$ slowly approximates $f$, providing a temporally consistent baseline.    
    A prediction is considered reliable if it agrees with $f^{te}$, i.e., $\argmax_{k} \hat{p}^{te}(r_{ui}) = \argmax_{k} \hat{p}(r_{ui})$.
\end{itemize}
\noindent
Figure \ref{fig:filtering} shows that temporal consistency is better suited for identifying accurate predictions with smaller errors in various training stages, ultimately leading to improved recommendation performance.
Further analysis revealed that the maximum probability varies significantly according to the \so-affinity of each unrated data.
The average maximum probability for the unrated data with the lowest 50\% affinity was over 20\% lower than that observed for the top 50\% affinity.
This large disparity can make maximum probability less effective as an indicator.
On the other hand, temporal consistency is not directly influenced by the magnitude of probability, making it more robust to varying affinities.
Based on these results, we employ temporal consistency as a reliability metric~for~filtering.
\textcolor{red}{For further analysis, refer to Appendix C.}

\smallsection{Self-distillation.}
Using reliable predictions identified through the filtering, we refine the model's preference knowledge.
Specifically, we train the model to make more confident preference predictions for the unrated data that it has reliably predicted previously.
This is achieved by reducing the entropy of the predicted distribution, which is a well-established technique for enhancing prediction confidence \cite{grandvalet2004semi}:
\begin{equation}
\begin{aligned}
\label{eq:RC}
    \ell_{sd} = \mathbbm{1}[\argmax_{k} \hat{p}^{te}(r_{ui}) = \argmax_{k} \hat{p}(r_{ui})] {H}(\hat{p}(r_{ui})),\\
\end{aligned}
\end{equation}
where $\mathbbm{1}[\cdot]$ is the indicator function and the entropy is defined as ${H}(\hat{p}(r_{ui})) = - \sum_k \hat{p}(r_{ui}=k) \log \hat{p}(r_{ui}=k)$.
By iterating this process, the model undergoes gradual refinement based on reliable predictions, acquiring preference knowledge for an increasing number of unrated data throughout the training.

\subsubsection{\textbf{Confidence-penalized preference distillation}}
\label{subsub:kd}
To facilitate more accurate preference discovery, we propose to use the biased teacher model.
As shown in Section \ref{subsec:anal}, the biased teacher can make accurate predictions for potential preference on high-affinity data.
We exploit these predictions as additional supervision for model training, supplementing limited rated data and improving its preference knowledge.
The biased teacher can be any recommendation model that is trained without debiasing learning.

However, naive training on high-affinity data can lead to reliance on specific patterns within the subset (e.g., high conformity to public opinion, as discussed in Section \ref{subsec:anal_setup}) which cannot be generalized to the remaining data.
Noticing this issue, we propose a new confidence-penalized distillation.
We combine the distillation loss with a confidence penalty, guiding our model to preserve the exact teacher prediction while preventing it from becoming overly confident.
\begin{equation}
\begin{aligned}
\label{eq:CP}
    \ell_{pd} = \lambda(\mathbb{E}_{\hat{p}}[r_{ui}] - t_{ui})^2 - {H}(\hat{p}(r_{ui}))\\
\end{aligned}
\end{equation}
The first term distills the biased teacher prediction $t_{ui}$ by compelling our model to produce an output distribution with the expected value of $t_{ui}$.
The second term applies the confidence penalty~that~discourages distributions with low entropy.
$\lambda$ is a hyperparameter to balance two terms.

\textcolor{red}{This confidence penalty is based on \textit{confidence regularization} for classifiers \cite{pereyra2017regularizing}, where overconfident outputs are known to lead to poor generalization.
In our context, it helps prevent the model from overfitting to the biased teacher’s overly certain outputs, particularly for high-affinity samples.
As will be shown in our ablation study (Section~\ref{subsubsec:ablation}), this penalty plays a crucial role in ensuring effective distillation.
Without it, the model tends to overfit to the teacher's biased predictions, resulting in significantly degraded accuracy on the counterfactual test.}

\subsubsection{\textbf{Soft and hard combination of dual distillations}}
\label{subsub:combine}
We balance two distinct distillation strategies based on the \so-affinity of each unrated data.
We introduce soft and hard combinations.
The soft combination integrates two strategies according to the predicted affinity: $a_{ui} = \hat{p}(s_{ui}=1)$, whereas the hard combination exclusively uses one strategy: $a_{ui}=1$ if $(u,i) \in \mathcal{S}^{01}$, otherwise $0$. 
\begin{equation}
\begin{aligned}
\label{eq:combine}
\small
\min_{\phi, \eta}\mathcal{L}_{T3} =|\mathcal{S}^0|^{-1}\sum_{(u,i)\in \mathcal{S}^{0}} a_{ui}\,\ell_{pd}(u,i) + (1-a_{ui})\ell_{sd}(u,i)
\end{aligned}
\end{equation}
\proposed is trained using either the soft or hard combination, which we call \underline{\proposed-Soft} and \underline{\proposed-Hard}, respectively.

\subsection{\textbf{The overall training objective}}
\label{subsec:method_train}
The final training loss of \proposed is as follows:
\begin{equation}
\begin{aligned}
    \min_{\phi, \eta} \max_{f_d} \mathcal{L}_{T1} + \alpha \mathcal{L}_{T2} + \beta \mathcal{L}_{T3},
\end{aligned}
\end{equation}
where $\alpha$ and $\beta$ are hyperparameters balancing the losses.
The \so-affinity is precomputed before the training, as explained in Section~\ref{subsec:s1affinity}.
Note that we only use the backbone model $f$ for inference, thus incurring no extra inference costs.
The training algorithm is presented in Appendix B.
\textcolor{red}{We also provide a detailed hyperparameter study in Section~\ref{subsub:hyper}.}

%% file: Sections/040Experiment.tex

\subsection{\textbf{Experiment setup}}
\label{Sec:experiements}

\smallsection{{Datasets.}}
We use three real-world datasets: Yahoo!R3 \cite{marlin2012collaborative}, Coat \cite{IPS}, and KuaiRec \cite{gao2022kuairec}.
They contain pre-split \textbf{{(a) biased training set}} from actual user-item interactions, and \textbf{{(b) counterfactual test set}}.\footnote{The counterfactual test data is obtained via RCT (Yahoo!R3, Coat) and full exposure of test items (KuaiRec). 
For KuaiRec, we convert the watch ratio to 1-5 rating scale to maintain consistency with other datasets \cite{gao2022kuairec}.}
As discussed in Section \ref{Sec:concept}, an ideal model for deployment should achieve high accuracy in both factual and counterfactual test environments \cite{InterD}.
For comprehensive evaluation in both tests environments, we randomly split out 10\% of the biased training set to form \textbf{{(c) factual test set}}.

\begin{table}[h]
\caption{Data statistics after preprocessing.}
\label{tab:statistics}
\resizebox{\linewidth}{!}{
\begin{tabular}{cccccc}
\toprule
\multicolumn{1}{l}{} & \multicolumn{1}{c}{\multirow{2}{*}{\#User}} & \multicolumn{1}{c}{\multirow{2}{*}{\#Item}} & \multicolumn{3}{c}{\#Ratings} \\ \cmidrule{4-6}
\multicolumn{1}{l}{} & \multicolumn{1}{c}{} & \multicolumn{1}{c}{} &  (a) training & (b) counterfactual test & (c) factual test\\ \midrule
Yahoo!R3 & 15,400 & 1,000 & 280,534 & 54,000 & 31,170 \\
Coat & 290 & 300 & 6,264 & 4,640 & 696 \\
KuaiRec & 7,176 & 10,728 & 11,277,725 & 4,676,570 & 1,253,081 \\
\bottomrule
\end{tabular}}
\end{table}

\smallsection{{Evaluation setup.}}
In this work, we specifically target multi-level explicit feedback where a special unbiased training set is not available, which is the identical setup as \cite{AT, saito2022towards}.
We employ two metrics widely used for rating prediction \cite{saito2022towards}: Mean Squared Error (MSE) and Mean Absolute Error (MAE). 
We hold out 10\% of the training set as the validation set.
We report the average value of five independent runs.
Note that we do not binarize ratings, as our focus is on multi-level feedback.

\smallsection{{Compared methods.}}
We focus on evaluating a model's capability to handle both factual and counterfactual tests simultaneously.
We exclude methods that require unbiased training data from the baselines \cite{autodebias, DRLTD, li2023balancing, liu2022kdcrec}, as we focus on scenarios without such data.

As our main competitors, we include recent methods which can be grouped into four relevant categories.
The first group follows \textbf{typical training} without any debiasing techniques.
\begin{itemize}
    \item \textbf{Standard Training} optimizes the model to minimize empirical risks on the collected feedback ($\mathcal{L}_{T1}$).
\end{itemize}
The second group includes advanced \textbf{adversarial learning}:
\begin{itemize}
    \item \textbf{FADA \cite{FADA}} proposes fine-grained adversarial learning for class-level feature alignment. It uses pseudo-labeling on the unlabeled target domain and aligns class-level distributions through a fine-grained discriminator.
    \item \textbf{IA \cite{FADAIA}} introduces an inference adjustment strategy to mitigate label distribution shift by correcting posterior probabilities. This adjustment is applied alongside FADA.
\end{itemize}
The third group includes \textbf{debiasing learning} methods. 
We compare \proposed with three recent methods that achieve highly competitive performance in the counterfactual test.
\begin{itemize}
    \item \textbf{Stable-DR \cite{STABLEDR}} proposes a stabilized doubly robust learning framework that cyclically updates imputation, propensity, and prediction models.
    \item \textbf{Stable-MRDR \cite{STABLEDR}} stabilizes an enhanced variant of doubly robust learning \cite{guo2021enhanced}, designed to reduce variance while preserving double robustness.
    \item \textbf{DCE-TDR \cite{DCETDR}} introduces a doubly calibrated estimator that calibrates both the imputation and propensity models.
\end{itemize}
The last group includes \textbf{knowledge distillation} methods:
\begin{itemize}
    \item \textbf{InterD \cite{InterD}} integrates knowledge from two teacher models—one biased and one debiased—by interpolating their predictions and distills the combined knowledge.
    \item \textbf{BPL} is our proposed method that employs dual distillation strategies: reliability-filtered self-distillation and confidence-penalized preference distillation. 
    We present results for two variants, \proposed-Soft and \proposed-Hard, which differ in their loss balancing strategies (Eq.\ref{eq:combine}).
\end{itemize}
For the \textbf{biased teacher} in both InterD and \proposed, we employ a recent model \cite{GLOCALK} that shows superior performance on the factual test.
For the debiased teacher in InterD, we utilize the best-performing debiasing method for each dataset.

\smallsection{{Implementation details.}} 
For backbone models, we employ the ID-based model \cite{koren2009matrix} for main experiments, following \cite{AT,saito2022towards,STABLEDR}.
We report results with other backbone models in Section \ref{subsub:backbone}.
We first tune the embedding size in $\{16, 32, 64, 128\}$ with the standard training on the validation set, and fix the size for all baselines.
The selected sizes are 64 (Yahoo!R3, KuaiRec), 32 (Coat), and 16 (Simulation).
Implementation details for baselines are provided in Appendix F.

We implement \proposed by using PyTorch with CUDA.
\textcolor{red}{All hyperparameters are tuned using grid search on the validation set.}
For affinity estimation, we train a simple binary classifier with binary cross-entropy loss (Section \ref{subsec:s1affinity}).
The classifier is a one-layer MLP with a sigmoid activation function.
We use Adam optimizer where the learning rate and weight decay are chosen from $\{10^{-6},..., 10^{-1}\}$.
We set $\lambda=1.0$, $\tau=0.999$, and tune $\alpha, \beta$ in $\{0.5, ..., 2.0\}$ and $x$ between $1$ and $30$.
\textcolor{red}{We provide a detailed hyperparameter study and guidelines for choosing values in Section~\ref{subsub:hyper}.}

\input{Sections/092main_dual}

\subsection{\textbf{Results on factual and counterfactual test environments}}
\label{subsec:dual}
We investigate the effectiveness of \proposed in both factual and counterfactual test environments.
Figure~\ref{fig:dual} presents the MSE and MAE results for both factual and counterfactual tests.
Furthermore, in Figure~\ref{fig:dual_h}, we report the harmonic mean of the scores on the factual and counterfactual tests to provide an overall comparison of a model’s capability \cite{InterD}.
For readability, we report the best performance from either FADA or FADA with IA (denoted as FADA/IA) and from either Stable-DR or Stable-MRDR (denoted as Stable-DR/MRDR).
\begin{itemize}
    \item The biased teacher shows superior performance in the factual test, but its performance is largely degraded in the counterfactual test.
    Conversely, debiasing learning (Stable-DR/MRDR, DCE-TDR) improves the performance on the counterfactual test, but it largely degrades performance on the factual test.
    This highlights the limited capability of the existing preference learning methods in minimizing errors across all user-item data.

    \item Advanced adversarial learning that performs class-level alignment (FADA/IA) achieve improvements to some extent. 
    They also cause less performance degradation on factual tests. 
    However, their effectiveness in counterfactual tests remains limited compared to debiasing~methods.
    
    \item InterD achieves a good balance between both tests by combining the knowledge from the biased and debiased teachers. 
    However, it often fails to bring large improvements compared to its debiased teacher (i.e., Stable-DR/MRDR).
    This shows the limitation of solely relying on fixed teacher knowledge for preference learning.
    
    \item \proposed achieves the best overall balance between both tests.
    It shows low errors in the factual test, and also the lowest errors in the counterfactual test.
    We attribute the performance gain to the proposed strategies that adaptively distill biased teacher knowledge and allow the model to continuously refine itself.

    \item \textcolor{red}{\textbf{BPL-hard vs. BPL-soft:} 
    Between the hard and soft combinations, the hard selection consistently yields lower errors across three different real-world datasets and achieves the lowest harmonic mean of errors on the factual and counterfactual tests.
    With the hard selection, the teacher's knowledge for data with low affinity is not used for distillation.
    This exclusive selection can be more beneficial when \so and \sz have a larger discrepancy.
    Moreover, the hard selection is more robust to the estimated affinity scores, as it does not use the individual values directly which will also be shown in Section \ref{exp:study_of_DSPL}.}
\end{itemize}


\subsection{\textbf{Study of \proposed}}
\label{exp:study_of_DSPL}

\subsubsection{\textbf{Ablation study}}
\label{subsubsec:ablation}
We provide a comprehensive ablation study to validate the effectiveness of each proposed component.
In Table \ref{tab:abl}, we compare six ablations of \proposed-Hard.
The first four ablations remove each loss: (1) $\mathcal{L}_{T2}$ (Eq.\ref{eq:Ld}),  (2) $\ell_{sd}$ (Eq.\ref{eq:RC}), (3) $\ell_{pd}$ (Eq.\ref{eq:CP}), and (4) $\mathcal{L}_{T2}$ and $\mathcal{L}_{T3}$.
Furthermore, we assess the effectiveness of two alternative design choices:
(5) `w/o confidence penalty' omits the penalty term from~$\ell_{pd}$, which can be thought of as the typical teacher-student distillation where the student is trained to mimic the teacher’s predictions. 
(6) `\so-affinity as propensity' reweights the training loss using the estimated affinity for each data, similar to propensity-based approaches \cite{IPS, STABLEDR}.

First, the best performance is achieved when using all losses, indicating that each contributes to improved performance on both factual and counterfactual tests. 
Notably, excluding either of the dual distillation losses (i.e., $\ell_{sd}$ and $\ell_{pd}$) causes a significant performance drop, highlighting that the dual distillation strategies effectively contribute complementary aspect of the preference learning.
\textcolor{red}{Second, the confidence penalty is crucial for preference distillation from the biased teacher. 
Removing the penalty results in a drastic performance drop on counterfactual tests across both datasets, indicating that the model overfits to the biased teacher’s predictions without the penalty.}
This shows the effectiveness of our penalty-combined distillation strategy, which discourage the model from learning patterns specific to high-affinity data. 
Moreover, reweighting the training loss using affinity scores proves highly ineffective.
\textcolor{red}{Lastly, we observe that aligning \sz and \so without considering $\mathcal{S}^{01}$ in Eq.\ref{eq:Ld} slightly degrades the final accuracy (1.043 on Yahoo!R3-RCT) and leads to slower convergence.}


\begin{table}[t]
\caption{Performance of various ablations of \proposed.}
\label{tab:abl}
\centering
\resizebox{0.95\linewidth}{!}{
\begin{tabular}{lcccc} \toprule
 & \multicolumn{2}{c}{Yahoo!R3} & \multicolumn{2}{c}{Coat} \\ \cmidrule(lr){2-3} \cmidrule(lr){4-5}
 & Factual & Counter- & Factual & Counter- \\
  &  & factual &  & factual \\\midrule
\textbf{BPL-Hard} & \textbf{1.469} & \textbf{0.991} & \textbf{1.104} & \textbf{1.037} \\ \midrule
\textbf{Learning objective} &  &  &  &   \\
w/o $\mathcal{L}_{T2}$ & 1.509 & 1.219 & 1.119 & 1.120 \\
w/o $\ell_{sd}$ ($\mathcal{L}_{T3}$) & 1.426 & 2.035 & 1.089 & 1.211 \\
w/o $\ell_{pd}$ ($\mathcal{L}_{T3}$) & 1.494 & 2.304 & 1.149 & 1.242 \\
w/o $\mathcal{L}_{T2}$ \& $\mathcal{L}_{T3}$ & 1.539 & 1.938 & 1.123 & 1.287 \\ \midrule
\textbf{Training technique} &  &  &  &  \\  
w/o confidence penalty & 1.454 & 2.258 & 1.101 & 1.371 \\
\so-affinity as propensity  & 1.720 & 1.694 & 1.294 & 1.192 \\ \bottomrule
\end{tabular}}
\end{table}





\begin{figure}[t]
    \centering
    \hspace{-0.1cm}
    \includegraphics[width=0.24\textwidth]{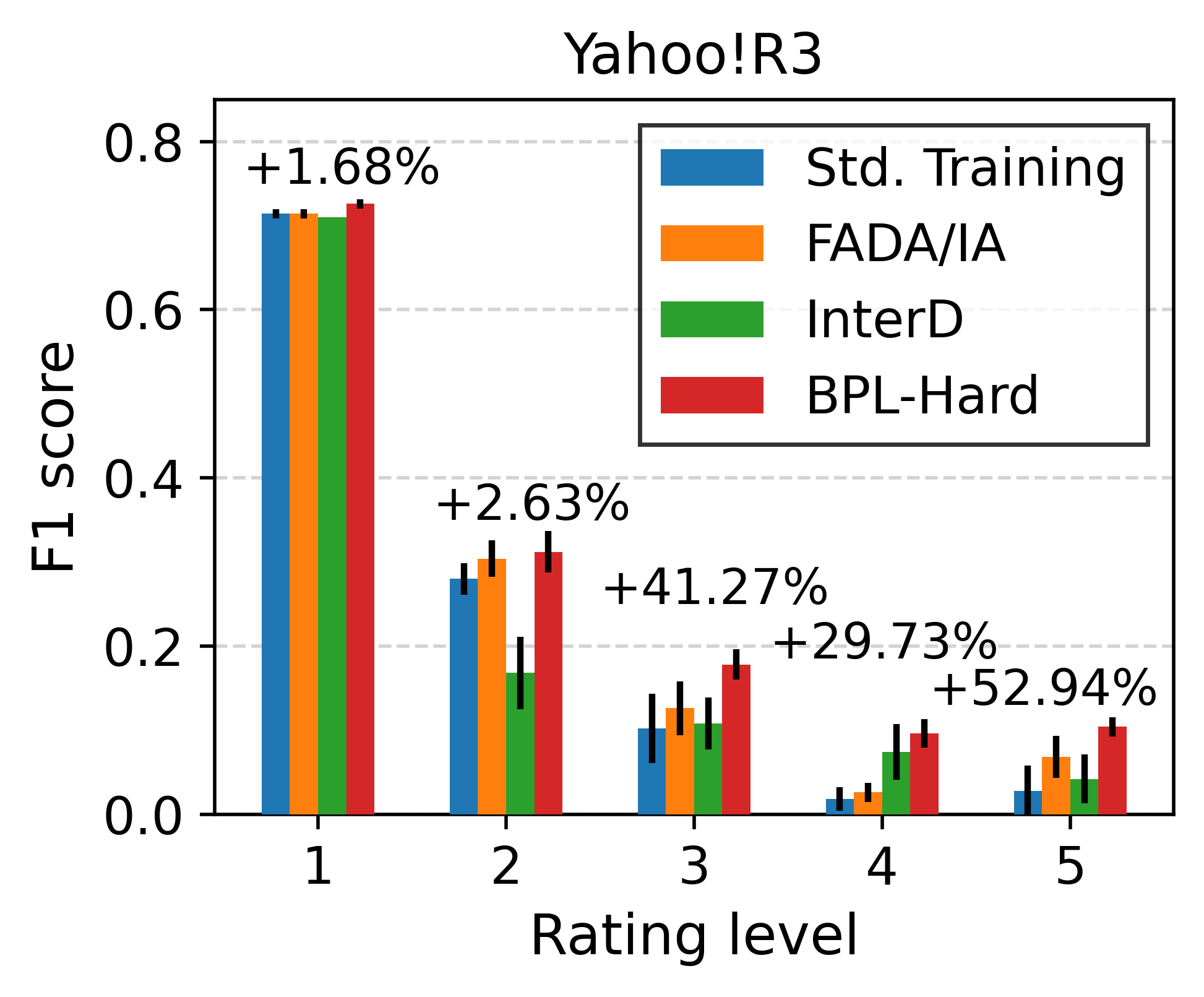}
    \hspace{-0.1cm}
    \includegraphics[width=0.24\textwidth]{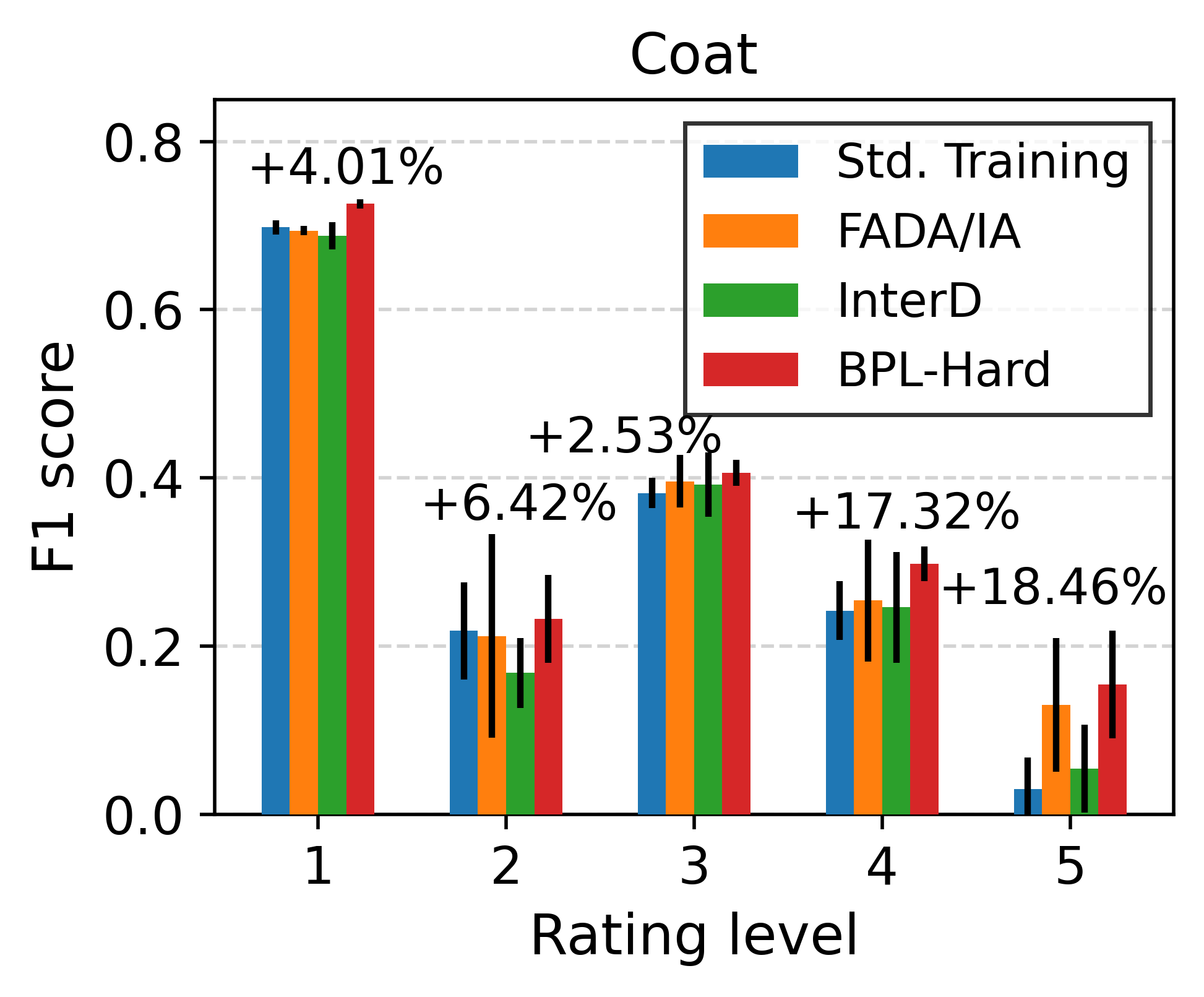}
    \caption{Evaluation of representation quality. We report the F1 score for each rating~level.} 
    \label{fig:F1}
    \vspace{-0.3cm}
\end{figure}


\subsubsection{\textbf{Representation quality analysis}}
We further analyze the quality of representations generated by various preference learning methods.
Specifically, we evaluate the rating discriminability of these representations—assessing whether they contain sufficient preference information to distinguish different ratings.
As discussed ealier, this measure is closely related to T3.
To perform this analysis, we conduct a rating-level classification using \textit{the fixed representations} generated by each method on the counterfactual test set.
We employ a two-layer softmax classifier and perform 5-fold cross-validation.

Figure \ref{fig:F1} shows that \proposed achieves the highest F1 scores, with particularly significant improvements on high ratings. 
The counterfactual test is generated through a random exposure process, which typically results in test items with low popularity and a tendency toward lower ratings.
This makes high ratings relatively rare and more challenging to predict.
This result show that \proposed indeed better encodes preference information into the representations, enabling clearer distinctions between different rating levels.
Furthermore, \proposed outperforms FADA/IA, which also employs adversarial learning strategies.
This highlights the superiority of our distillation strategies in capturing underlying preference on unrated data.

\subsubsection{\textbf{Sensitivity to affinity estimation}}
\label{subsec:sensitivity}
\begin{figure}[t]
    \centering
    \includegraphics[height=3.5cm]{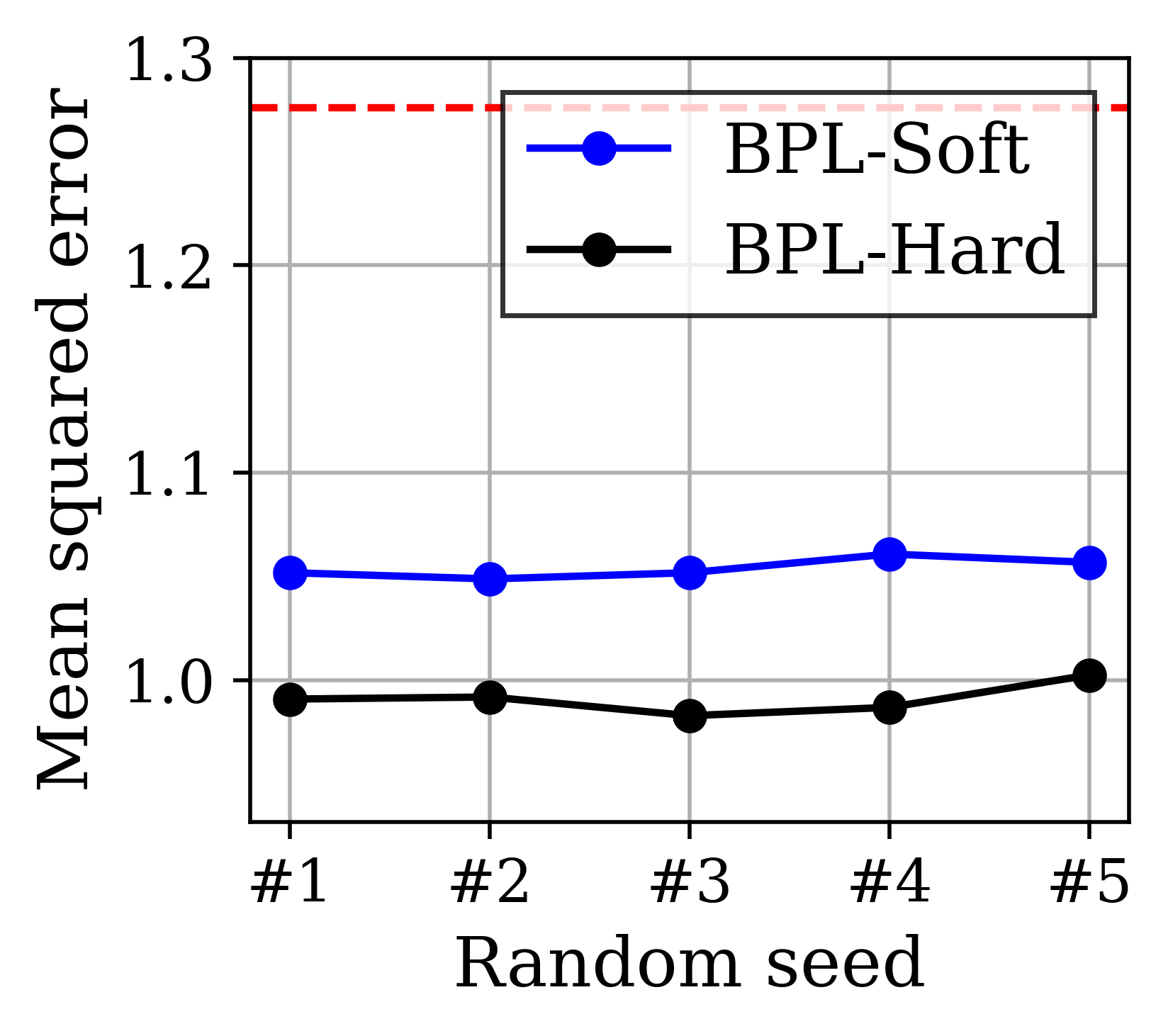}
    \hspace{0.1cm}
    \includegraphics[height=3.5cm]{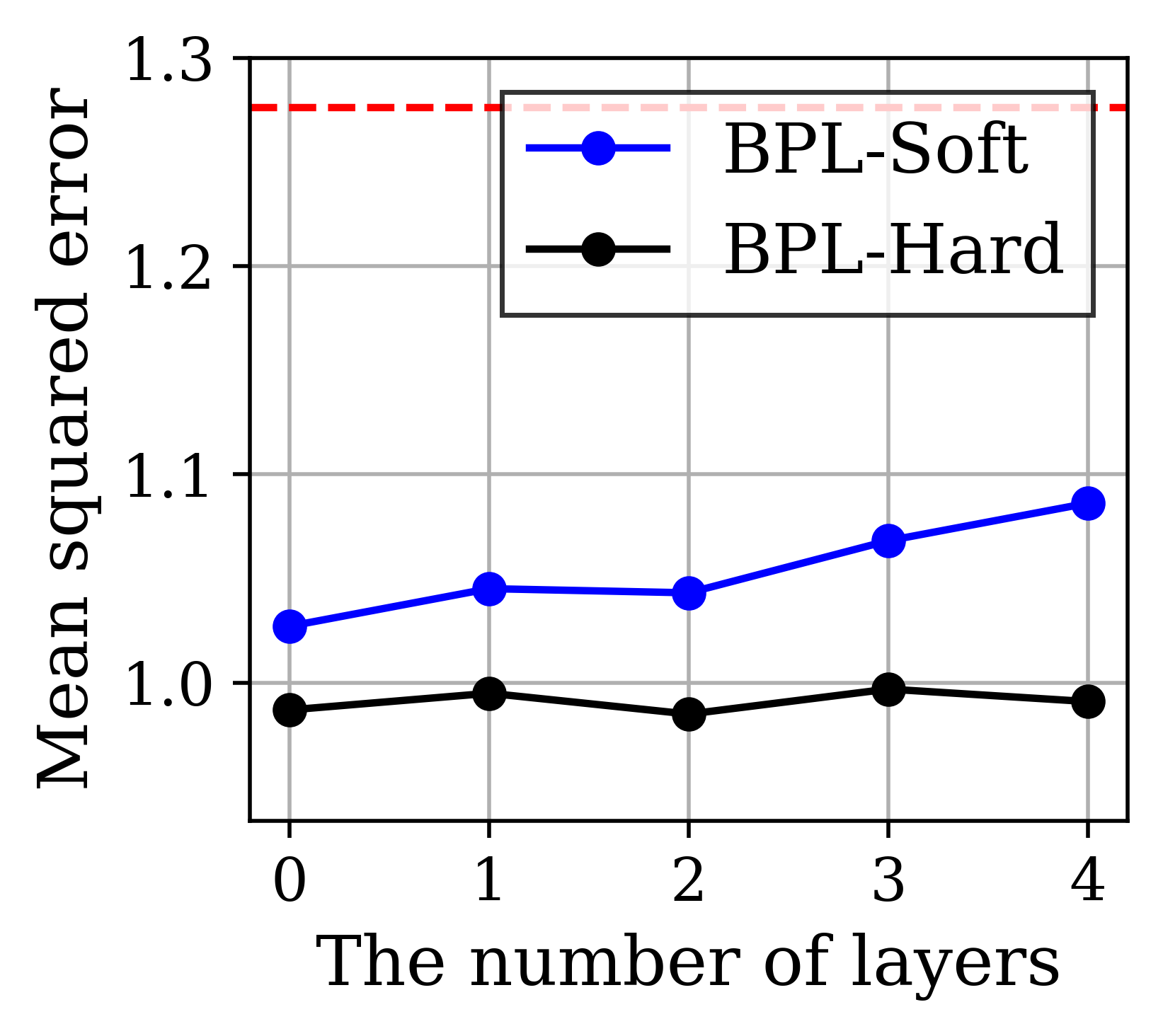}
    \caption{Results with varying \so-affinity estimation models. The red dotted line indicates the best competitor,~i.e.,~InterD.} 
    \label{fig:sensitivity}
    \vspace{-0.1cm}
\end{figure}

\proposed utilizes \so-affinity of unrated data obtained by a simple binary classifier (Section \ref{subsec:s1affinity}).
Here, we investigate how varying affinity estimation models influence \proposed.
We create five variants by adjusting random seeds, and another five variants by modifying the number of layers. In our experiments, we employ the single-layer model.
Figure \ref{fig:sensitivity} shows results on the counterfactual test of Yahoo!R3.
\textcolor{red}{Both \proposed-Soft and \proposed-Hard show a considerable degree of robustness and largely outperform the best competitor (i.e., InterD) in all setups.
Notably, \proposed-Hard shows highly stable performance with all 10 estimation models.
\proposed-Hard can have such robustness because it uses the estimated affinities only for finding a small set of high-affinity data, without using the individual values.
}

\begin{table}[t]
\caption{MSE results with varying hyperparameters. H-mean refers to the harmonic mean of the two scores.}
\label{tab:hyperparameter}
\centering
\resizebox{0.99\linewidth}{!}{%
\begin{tabular}{ccccccccc}
\toprule
\multicolumn{1}{l}{} & \multicolumn{4}{c}{Yahoo!R3} & \multicolumn{4}{c}{Coat} \\
\cmidrule(lr){2-5} \cmidrule(lr){6-9}
 &  & \multirow{2}{*}{Factual}  & Counter- & \multirow{2}{*}{H-mean} &  & \multirow{2}{*}{Factual} & Counter- & \multirow{2}{*}{H-mean} \\
 &  &  & factual &  &  &  &factual &  \\
\midrule
\multirow{3}{*}{$\alpha$} & 0.5 & 1.469 & 0.991 & 1.184 & 0.5 & 1.104 & 1.037 & 1.069 \\
 & 0.1 & 1.465 & 1.016 & 1.200 & 0.1 & 1.109 & 1.041 & 1.074 \\
 & 0.01 & 1.472 & 1.020 & 1.205 & 0.01 & 1.098 & 1.038 & 1.067 \\
\midrule
\multirow{4}{*}{$\beta$} & 2.0 & 1.471 & 0.986 & 1.181 & 2.0 & 1.115 & 1.045 & 1.079 \\
 & 1.5 & 1.469 & 0.991 & 1.184 & 1.5 & 1.115 & 1.046 & 1.079 \\
 & 1.0 & 1.478 & 1.090 & 1.255 & 1.0 & 1.104 & 1.037 & 1.069 \\
 & 0.5 & 1.484 & 1.420 & 1.451 & 0.5 & 1.138 & 1.161 & 1.149 \\
\midrule
\multirow{3}{*}{$x$} & 1\% & 1.469 & 0.991 & 1.184 & 20\% & 1.108 & 1.043 & 1.075 \\
 & 5\% & 1.480 & 1.037 & 1.220 & 25\% & 1.104 & 1.037 & 1.069 \\
 & 10\% & 1.505 & 1.051 & 1.238 & 30\% & 1.109 & 1.044 & 1.076 \\
\bottomrule
\end{tabular}}
\end{table}

\subsubsection{\textbf{Hyperparameter study}}
\label{subsub:hyper}
Table \ref{tab:hyperparameter} presents results of \proposed-Hard with varying hyperparameters.
Overall, \proposed shows a relatively stable performance on the factual test.
We mainly analyze the results in terms of counterfactual test results.

\begin{itemize}
    \item \textcolor{red}{For $\alpha$ and $\beta$, which control the effects of $\mathcal{L}_{T2}$ and $\mathcal{L}_{T3}$, the best performances are observed around 1 to 2 ($\alpha$) and around $0.5$ ($\beta$) on both datasets.}

    \item \textcolor{red}{For $x$, which controls the size of $\mathcal{S}^{01}$, it shows different tendencies for each dataset.
    Interestingly, we discover that its optimal value is closely related to the overall divergence between the training and test set.
    Specifically, Yahoo!R3 shows a higher divergence than Coat \cite{AT}; the KL-divergence of the rating distributions are 0.470 (Yahoo! R3) and 0.049 (Coat).
    As a result, smaller size of $\mathcal{S}^{01}$ is more beneficial for Yahoo!R3, whereas a relatively larger $\mathcal{S}^{01}$ can be used for Coat.
    This result suggests that $x$ can be selected considering the overall \so-affinity degree. 
    That is, datasets with higher divergence between observed and unobserved data may benefit from a more conservative (i.e., smaller) selection of $\mathcal{S}^{01}$.}
\end{itemize}



\subsubsection{\textbf{Results with other backbone models}}
\label{subsub:backbone}
\begin{figure}[t]
\centering
\begin{subfigure}{0.5\linewidth}
    \centering
    \includegraphics[height=3.3cm]{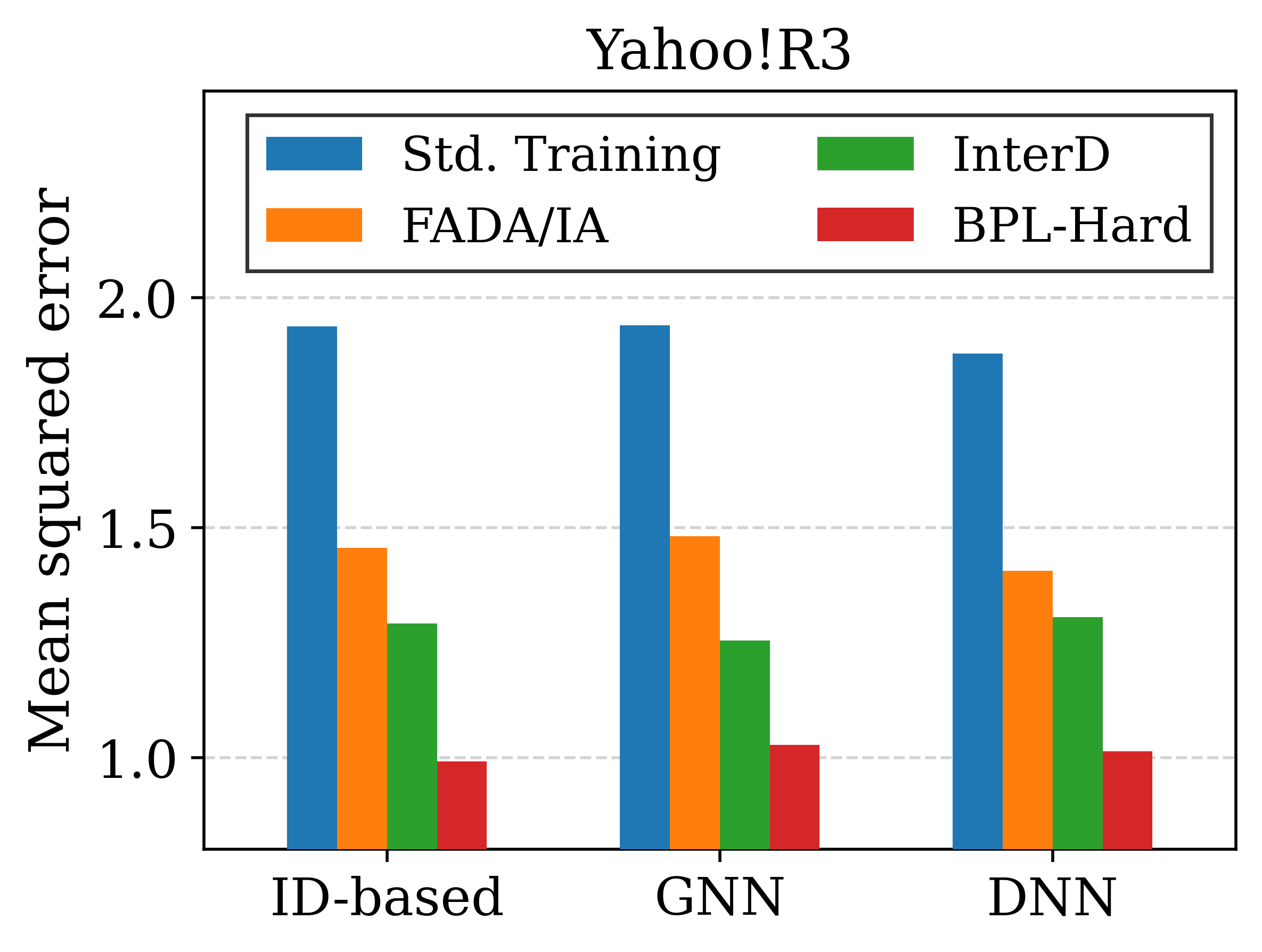}
\end{subfigure}%
\begin{subfigure}{0.5\linewidth}
    \centering
    \includegraphics[height=3.3cm]{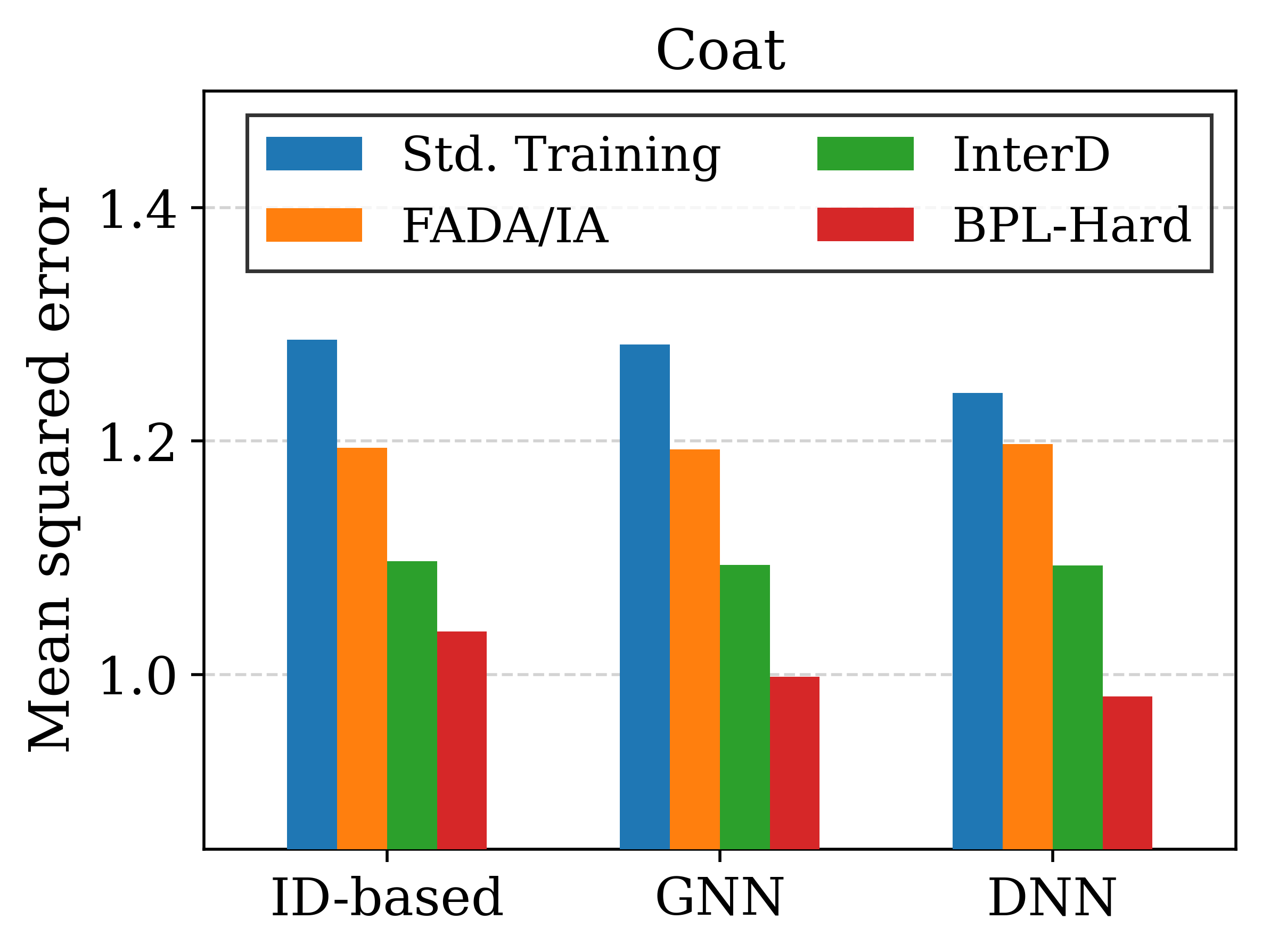}
\end{subfigure}\\
\caption{Results with various backbone models. 
We perform a paired t-test with the best competitor at the 0.05 significance level. 
\proposed-Hard achieves statistically significant improvements in six out of six cases.}
\label{fig:backbone}
\vspace{-0.4cm}
\end{figure}

\proposed can be flexibly applied to various recommendation models.
Figure \ref{fig:backbone} presents MSE results in the counterfactual test environment using various backbone models, including ID-based, graph neural network (GNN)-based \cite{berg2017graph}, and deep neural network (DNN)-based \cite{NCF} models, as done in \cite{STABLEDR, zhang2023invariant, li2022multiple}.
We compare baselines that show competitive performance in the previous experiments. 
We observe similar tendencies across the backbone models, which are consistent observations with the previous work \cite{STABLEDR, zhang2023invariant, li2022multiple}.


%



\subsection{\textbf{Results on benchmark counterfactual (RCT) test}} 
\label{subsec:RCT_result}
In the previous literature \cite{IPS, AT, saito2022towards, DRLTD}, RCT tests of Yahoo!R3 and Coat, have been used as benchmark evaluations.
To enable direct comparison under this setup, we report results on the RCT tests as well.
Note that the training set used in this section is \textit{larger} than that in previous experiments, as we use the \textbf{original training set} without splitting it for a factual test set, ensuring consistency with prior research.

\smallsection{{\textbf{Compared methods.}}}
Under RCT setup, we compare \proposed with various recent methods from related research fields.
Further details of baselines are provided in Appendix E.
\begin{itemize}
    \item \textbf{Adversarial \& semi-supervised learning}: {ADV} \cite{ganin2016domain}, {FADA} \cite{FADA}, {FADA/IA} \cite{FADAIA}, {FreeMatch} \cite{freematch}.

    \item \textbf{Debiasing learning for explicit feedback}:
    {IPS} \cite{IPS}, {AST} \cite{xiao2023reconsidering}, {AT} \cite{AT}, {DAMF} \cite{saito2022towards}, {MR} \cite{li2022multiple}, {TMRDR-CL} \cite{li2022tdr}, {Stable-DR} \cite{STABLEDR}, {Stable-MRDR} \cite{STABLEDR}, {DCE-TDR} \cite{DCETDR}.

    \item \textbf{Debiasing learning for other behavior types}: {TIDE} \cite{zhao2022popularity}, {Zerosum} \cite{Zerosum}, {ESAM} \cite{chen2020esam}, {InvCF} \cite{zhang2023invariant}.

    \item \textbf{Knowledge distillation}: {InterD} \cite{InterD}, {DebiasKD}, {\proposed}.
    DebiasKD is a new baseline that combines KD from the biased teacher with the best-performing debiasing method.
    It uses the biased teacher’s predictions for \szo (the same as \proposed-Hard) as additional supervision.
\end{itemize}

From Table~\ref{tab:main_table_b}, we have following observations:
First, the basic adversarial learning (ADV) falls short of effectively reducing errors.
Also, advanced methods to learn class-discriminative representations (FADA/IA) show limited effectiveness.
Similarly, FreeMatch based on semi-supervised learning fails to outperform debiasing methods. 
These results show the difficulty of capturing user preference from the biased and limited training data.
Among the debiasing baselines, methods designed for explicit feedback show competitive performance overall.
Also, Stable-DR/MRDR, based on the doubly robust approach, consistently shows~superior~performance.

KD-based baselines (i.e., InterD, DebiasKD) generally show highly competitive performance.
However, DebiasKD performs worse than the best debiasing method, indicating that a naive distillation from the biased teacher can negatively impact preference learning.
Conversely, InterD utilizes the prediction interpolation to control the impact of each teacher, outperforming DebiasKD and showing high effectiveness overall.
However, as it resorts to the pre-trained teachers, it fails to bring significant improvements compared to the debiased teacher.
Lastly, consistent with previous results, \proposed generally achieves the lowest errors among all compared~methods.

\input{Sections/091main_RCT}

%% file: Sections/092main_dual.tex
\begin{figure*}[t]
\centering
\begin{subfigure}[t]{0.15\linewidth}
    \includegraphics[width=\linewidth]{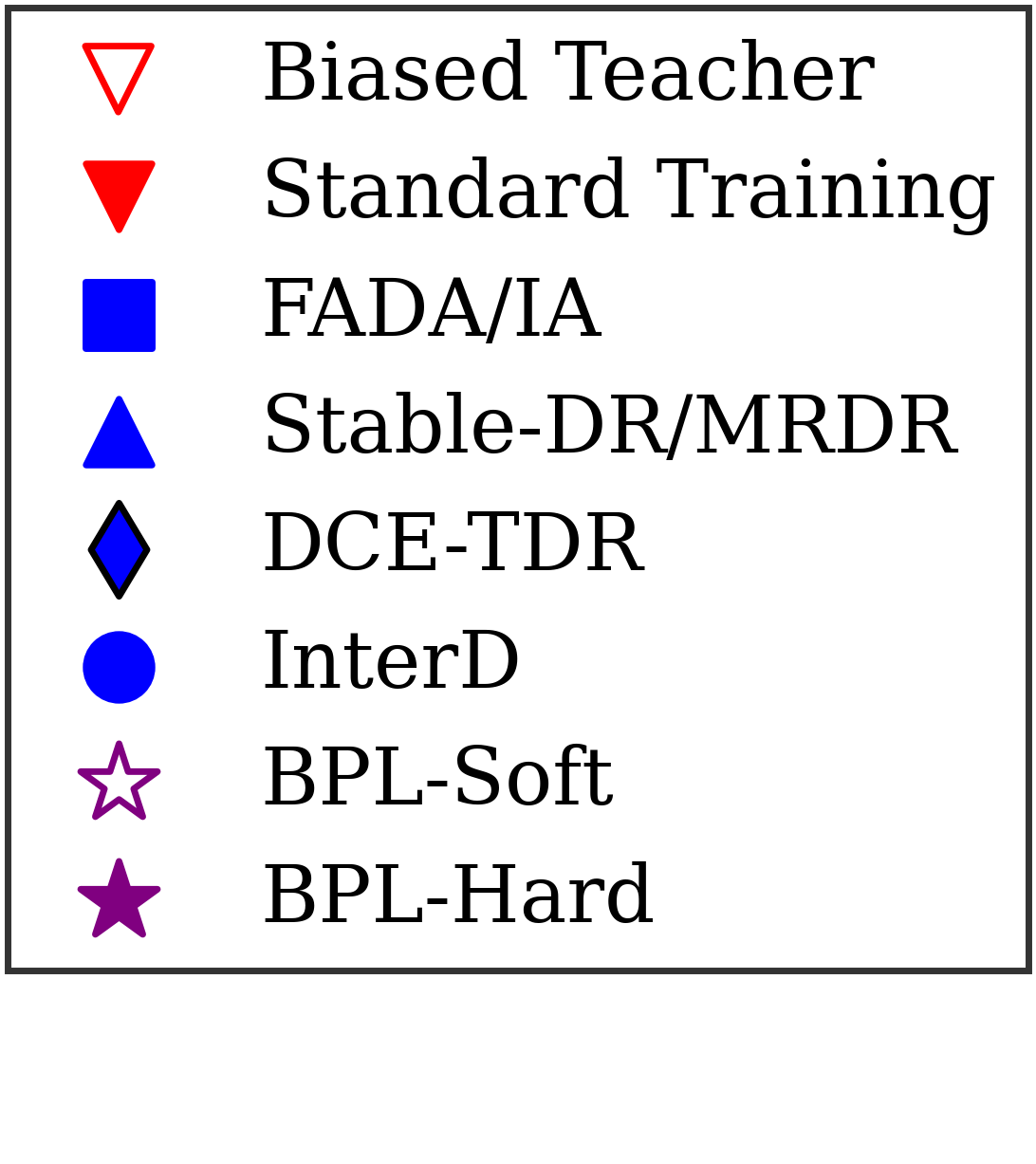}
\end{subfigure}
\hspace{0.1cm}
\begin{subfigure}[t]{0.28\linewidth}
    \includegraphics[width=\linewidth]{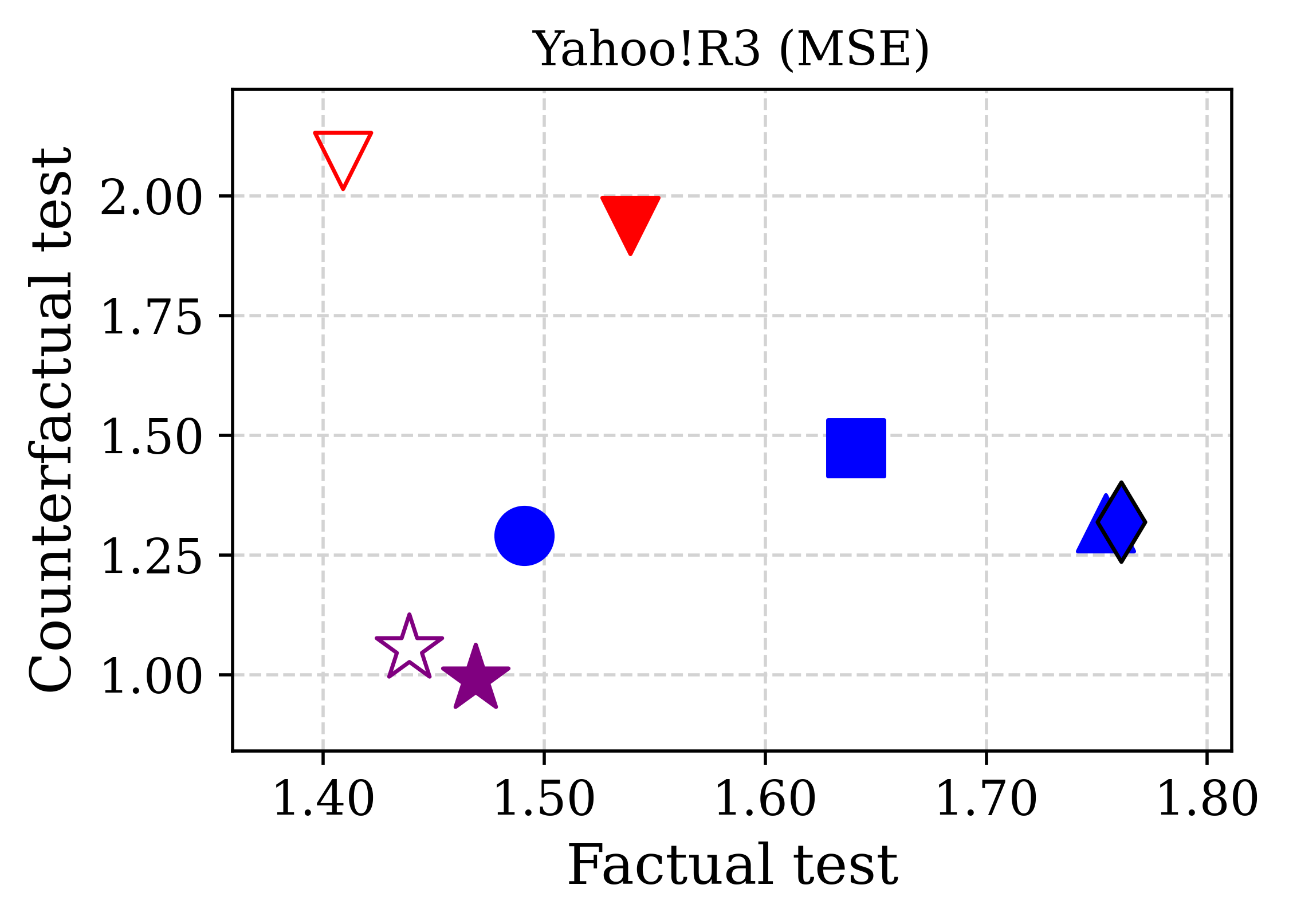}
\end{subfigure}
\begin{subfigure}[t]{0.27\linewidth}
    \includegraphics[width=\linewidth]{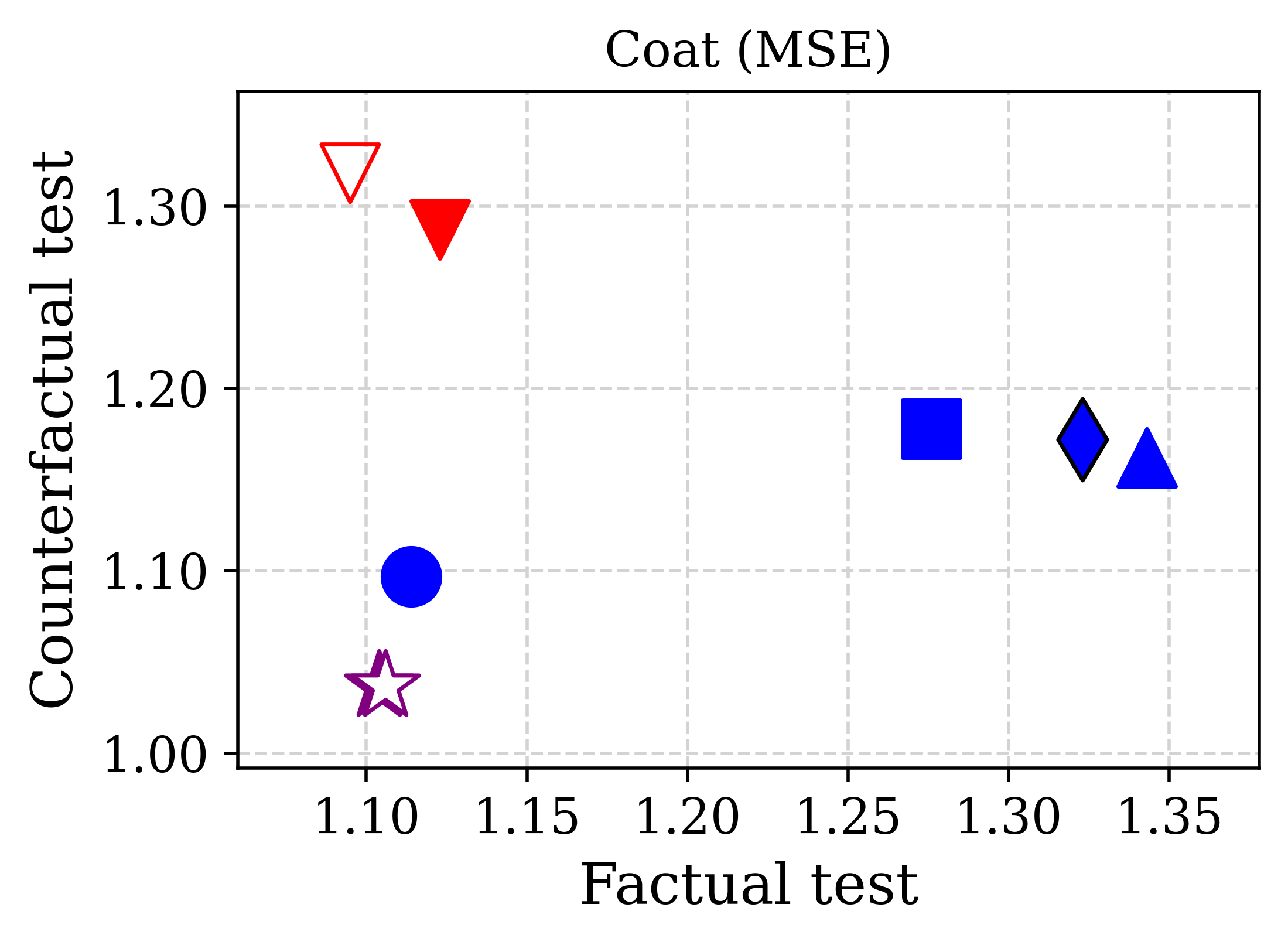}
\end{subfigure} 
\begin{subfigure}[t]{0.27\linewidth}
    \includegraphics[width=\linewidth]{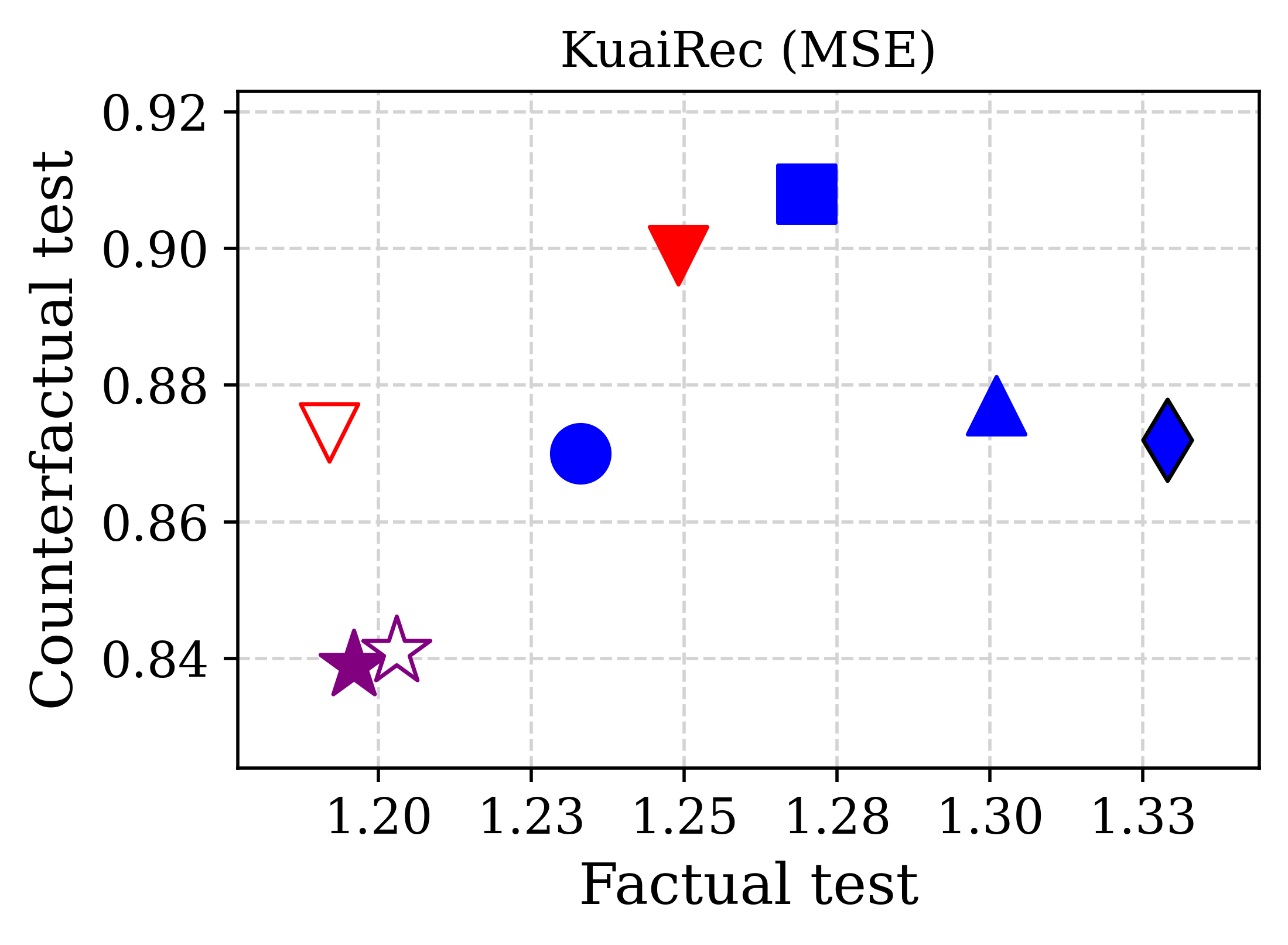}
\end{subfigure}
\vspace{0.2cm}
\begin{subfigure}[t]{0.15\linewidth}
    \includegraphics[width=\linewidth]{Figures/dual_legend.png}
\end{subfigure}
\hspace{0.1cm}
\begin{subfigure}[t]{0.275\linewidth}
    \includegraphics[width=\linewidth]{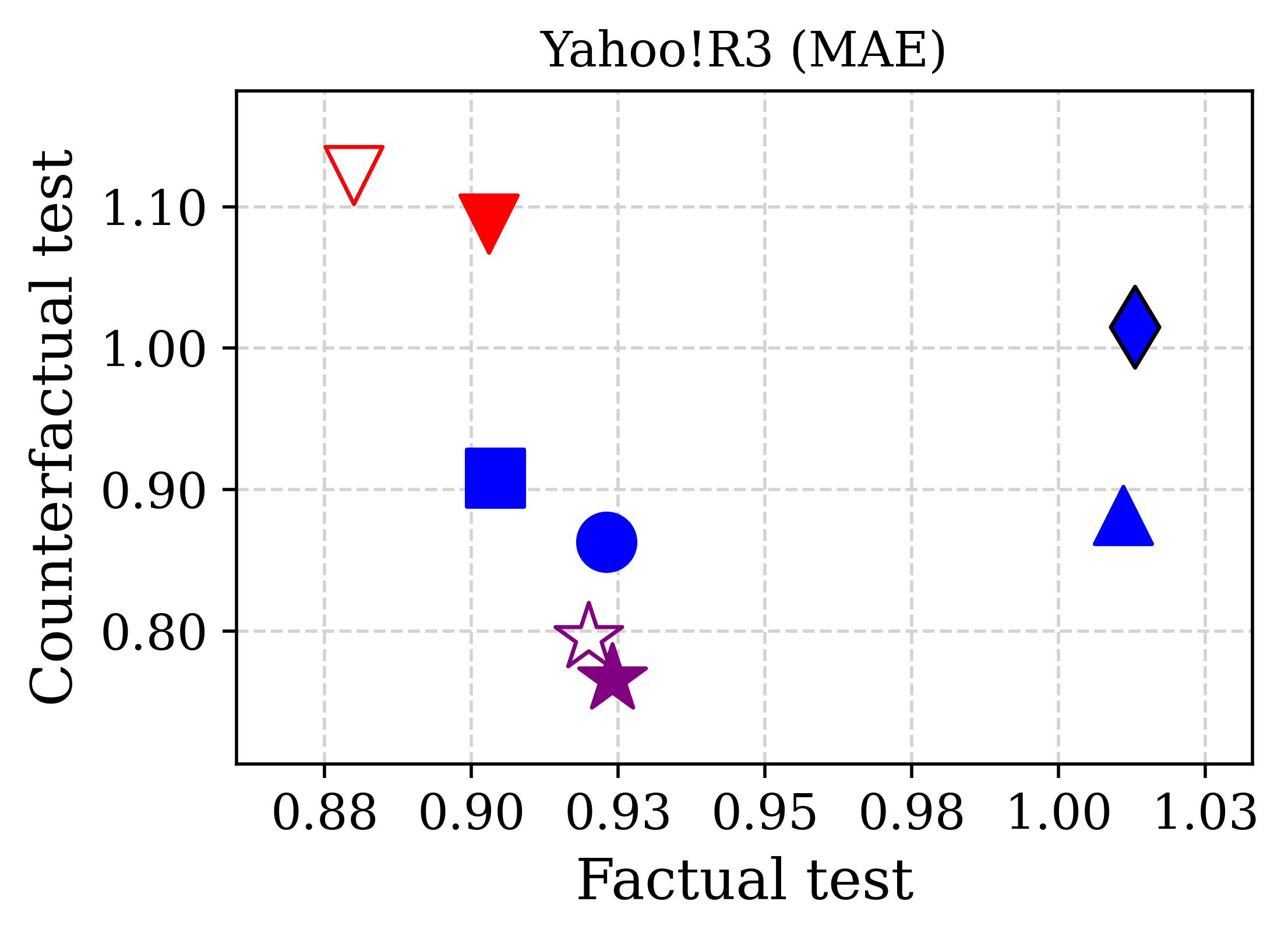}
\end{subfigure}
\begin{subfigure}[t]{0.27\linewidth}
    \includegraphics[width=\linewidth]{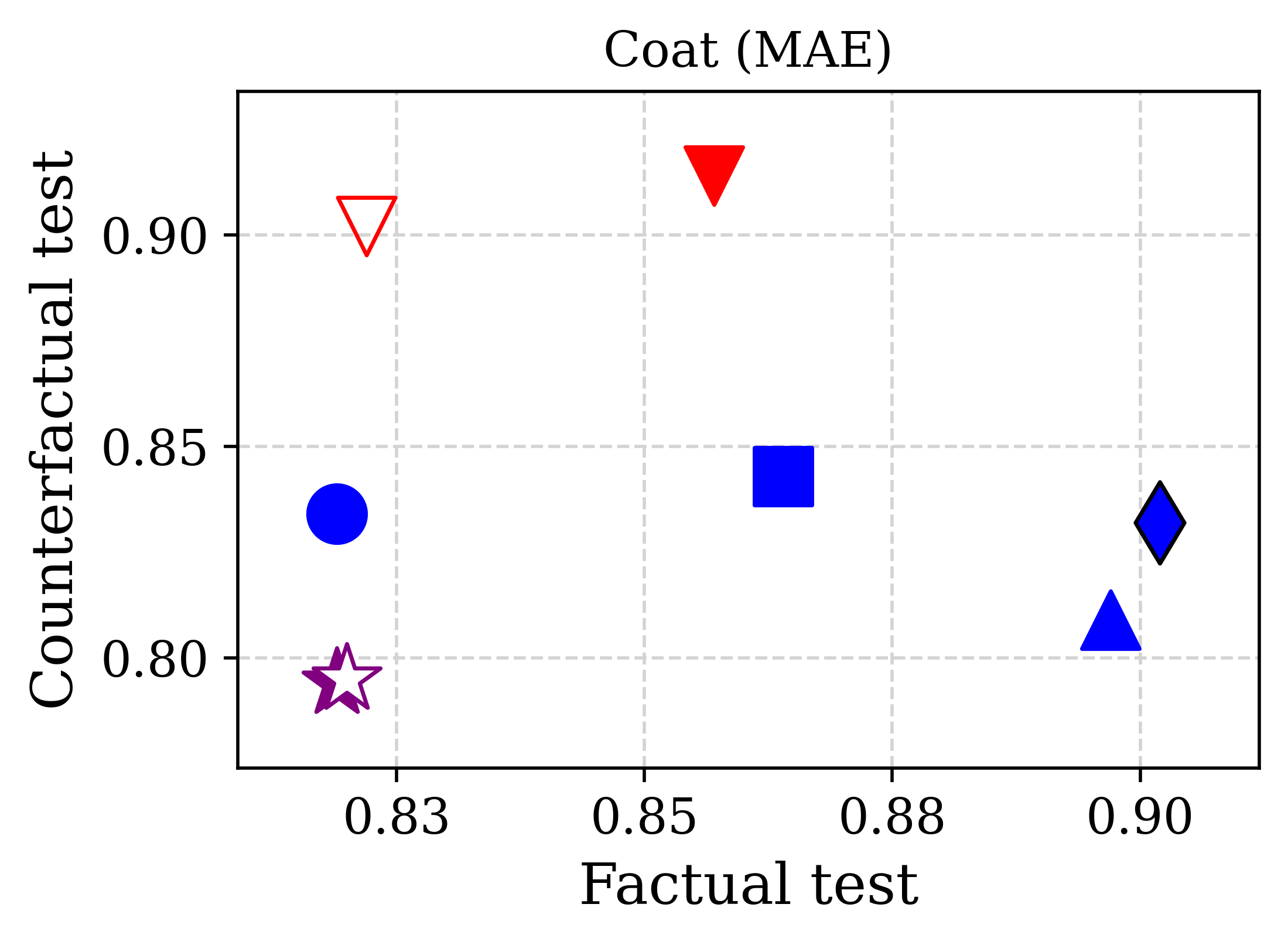}
\end{subfigure} 
\begin{subfigure}[t]{0.27\linewidth}
    \includegraphics[width=\linewidth]{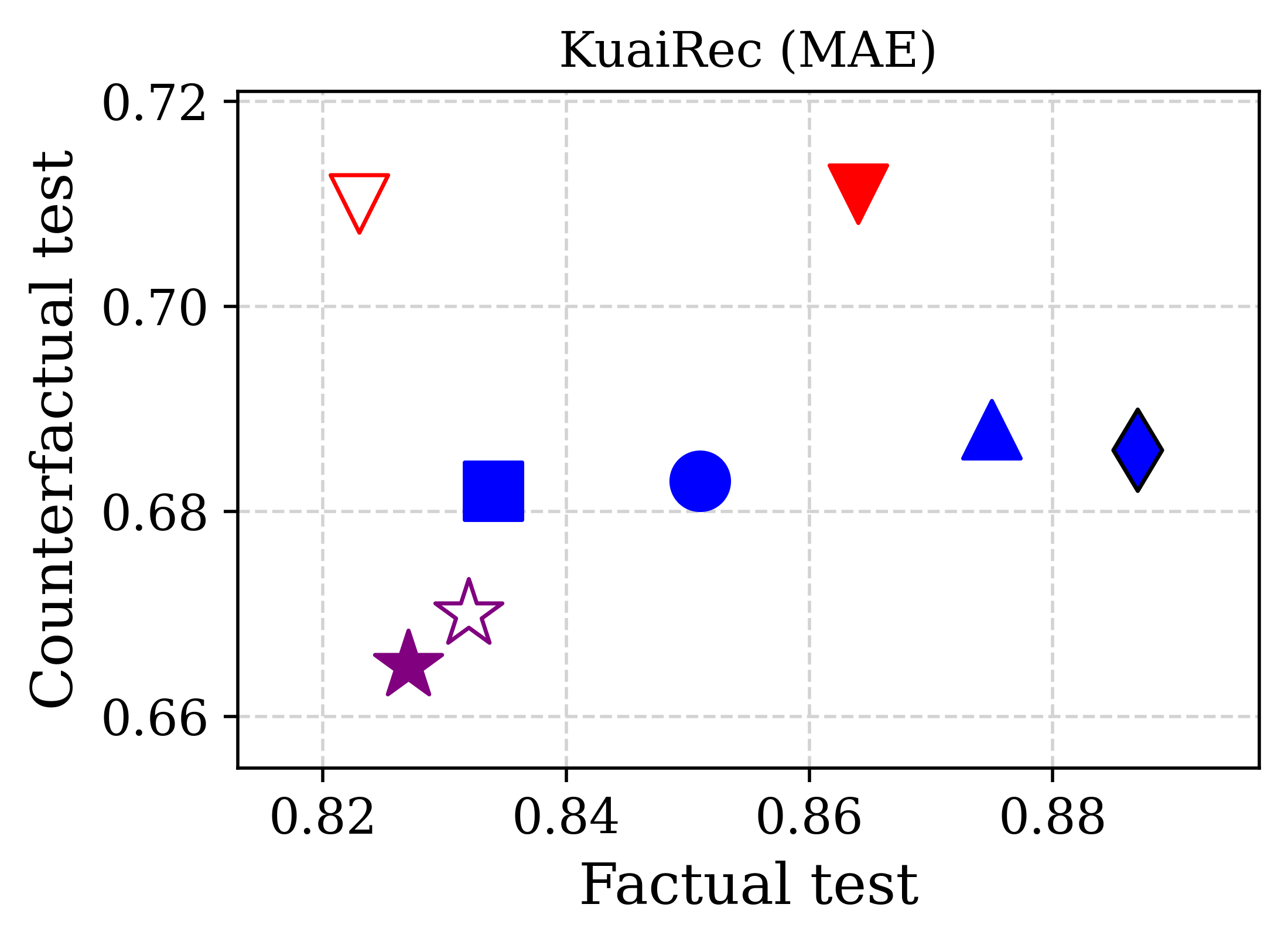}
\end{subfigure} 
\caption{Factual and counterfactual test results across three real-world datasets. We perform a paired t-test with the best competitor at the 0.05 significance level. 
\proposed-Soft achieves statistically significant improvements in four out of six cases, and \proposed-Hard significantly outperforms in five out of six cases on the counterfactual test.}
\label{fig:dual}
\vspace{-0.3cm}
\end{figure*}


\begin{figure*}[t]
\centering
\begin{subfigure}[t]{0.15\linewidth}
    \includegraphics[width=\linewidth]{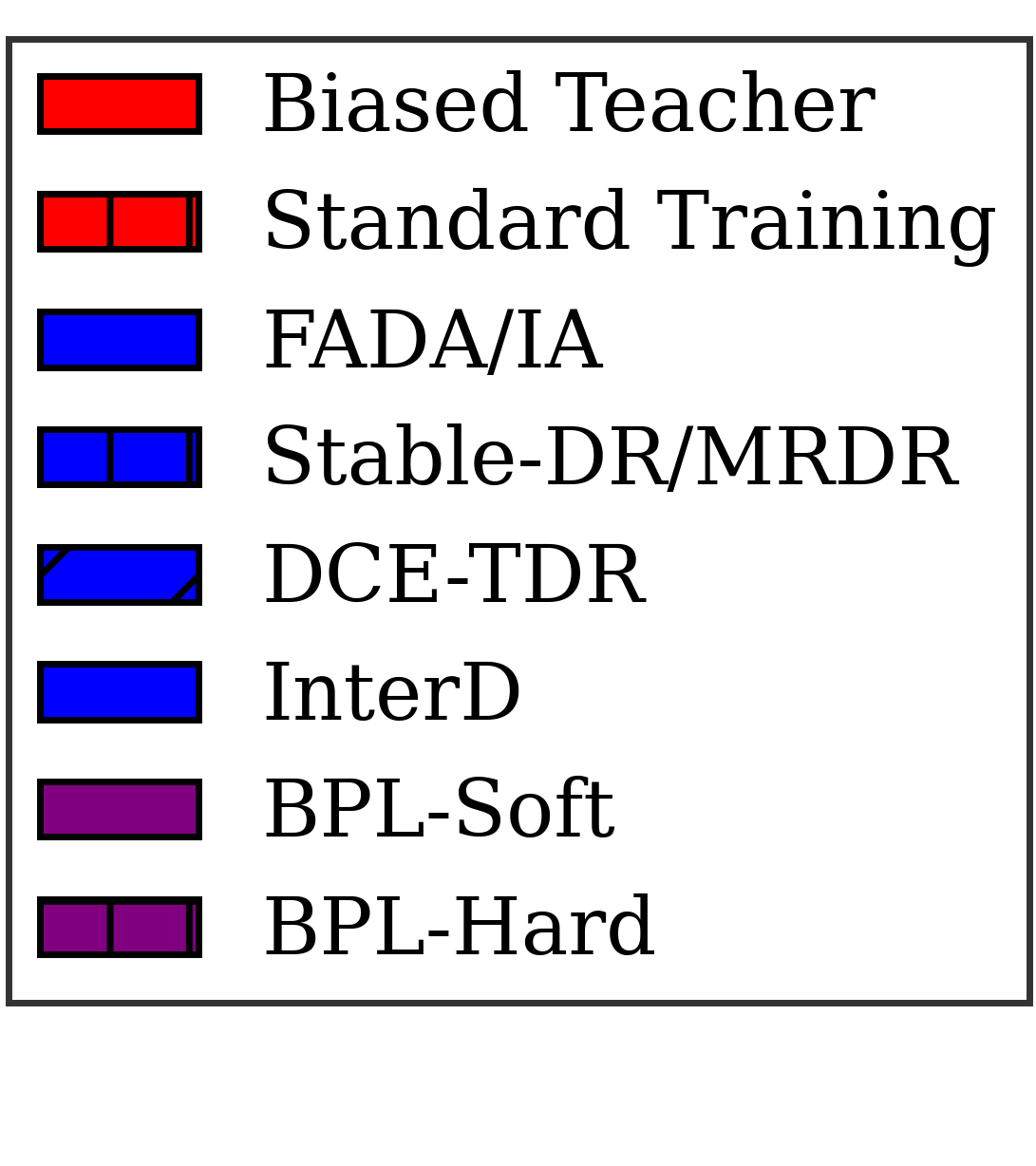}
\end{subfigure}
\hspace{0.1cm}
\begin{subfigure}[t]{0.27\linewidth}
    \includegraphics[width=\linewidth]{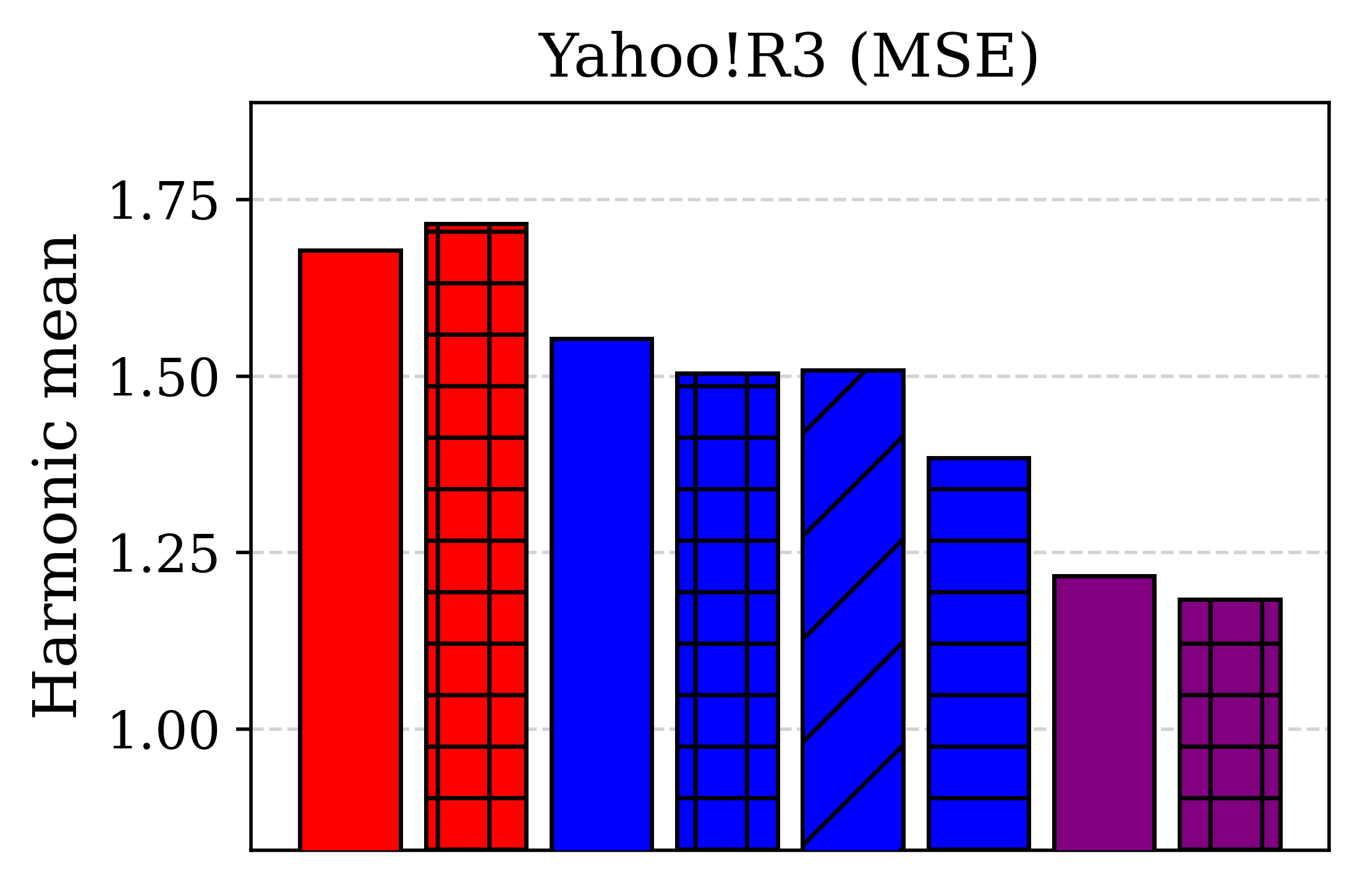}
\end{subfigure}
\begin{subfigure}[t]{0.27\linewidth}
    \includegraphics[width=\linewidth]{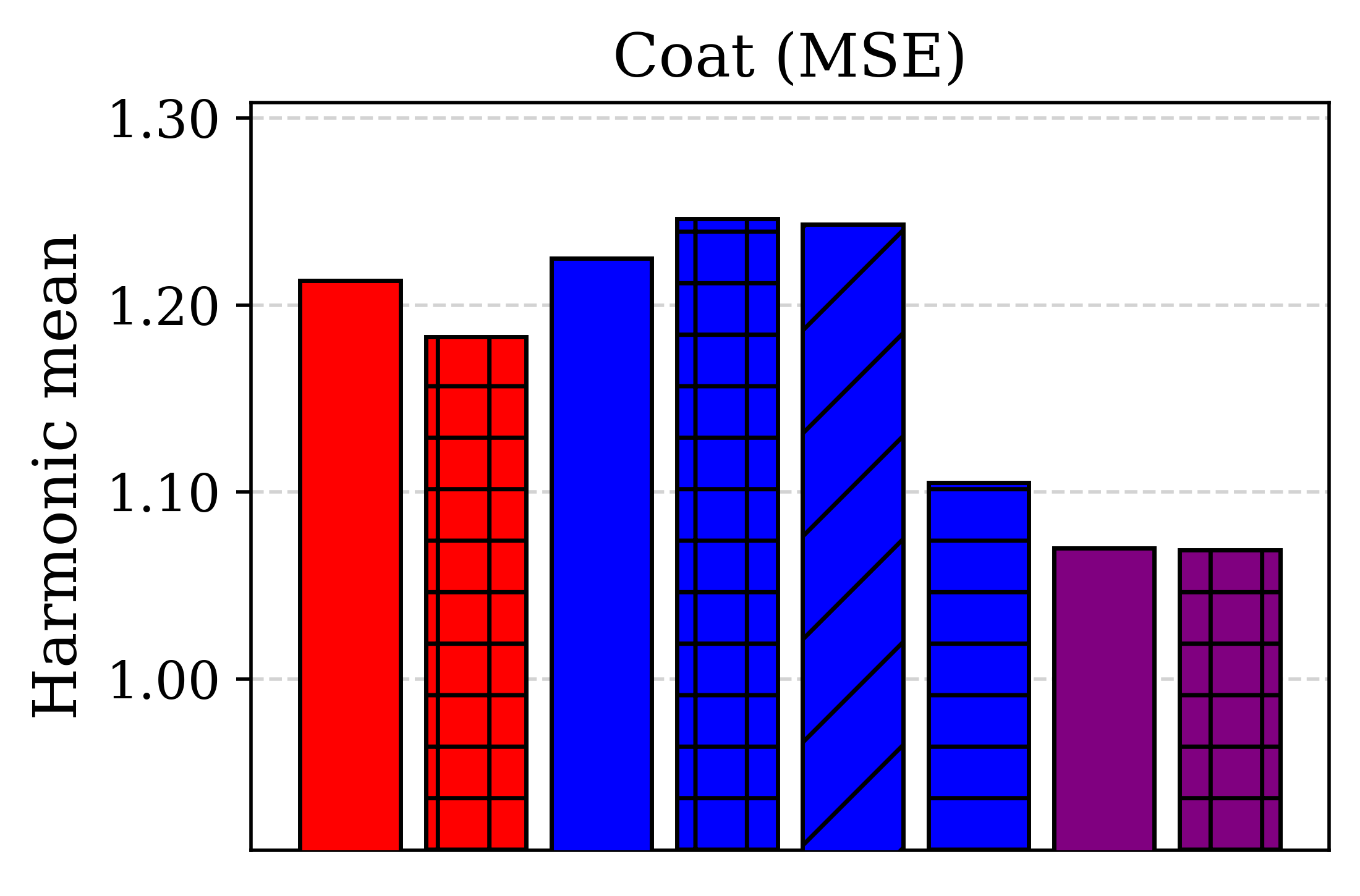}
\end{subfigure} 
\begin{subfigure}[t]{0.27\linewidth}
    \includegraphics[width=\linewidth]{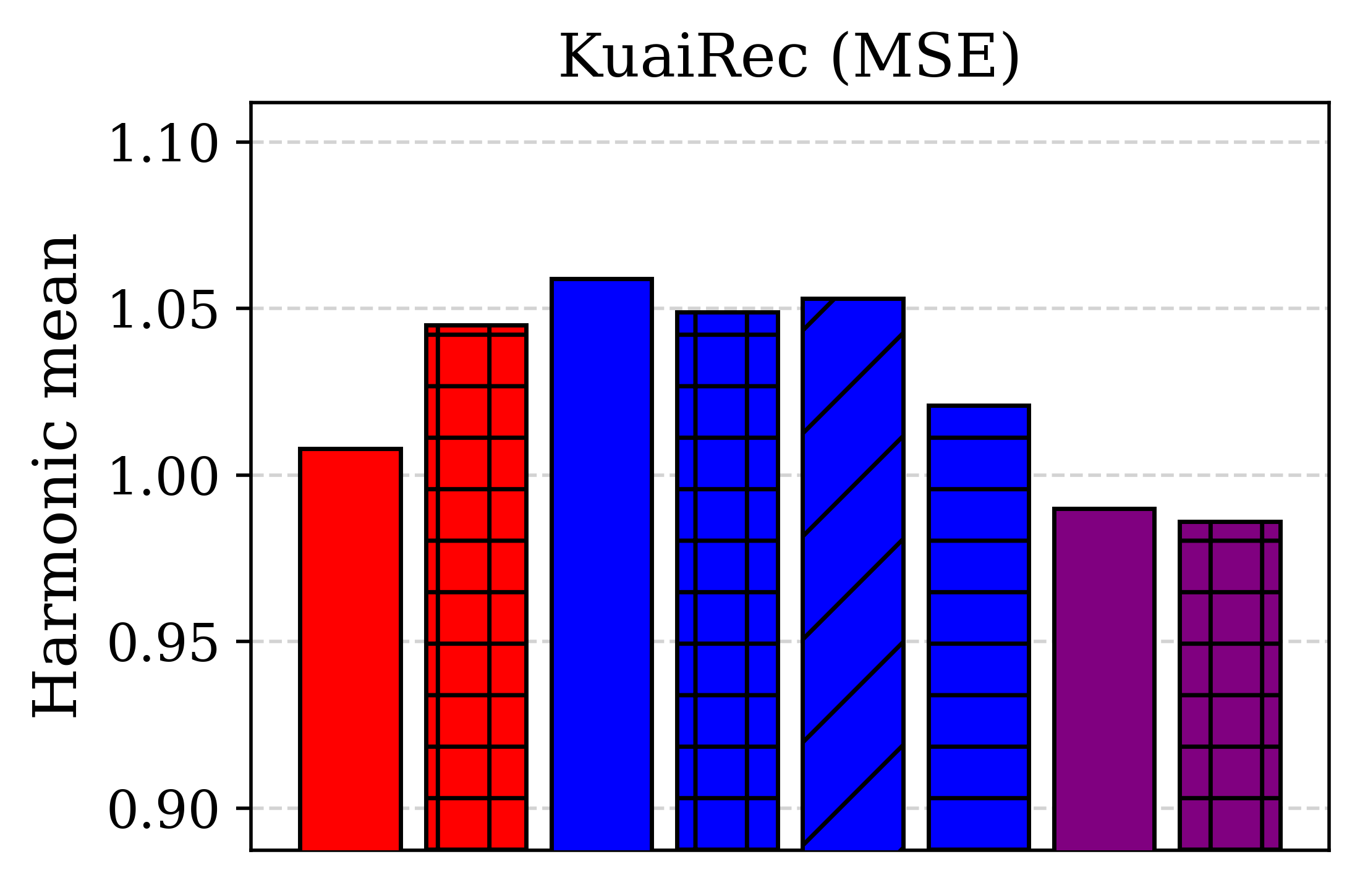}
\end{subfigure}
\vspace{0.2cm}
\begin{subfigure}[t]{0.15\linewidth}
    \includegraphics[width=\linewidth]{Figures/hmean_legend.png}
\end{subfigure}
\hspace{0.1cm}
\begin{subfigure}[t]{0.27\linewidth}
    \includegraphics[width=\linewidth]{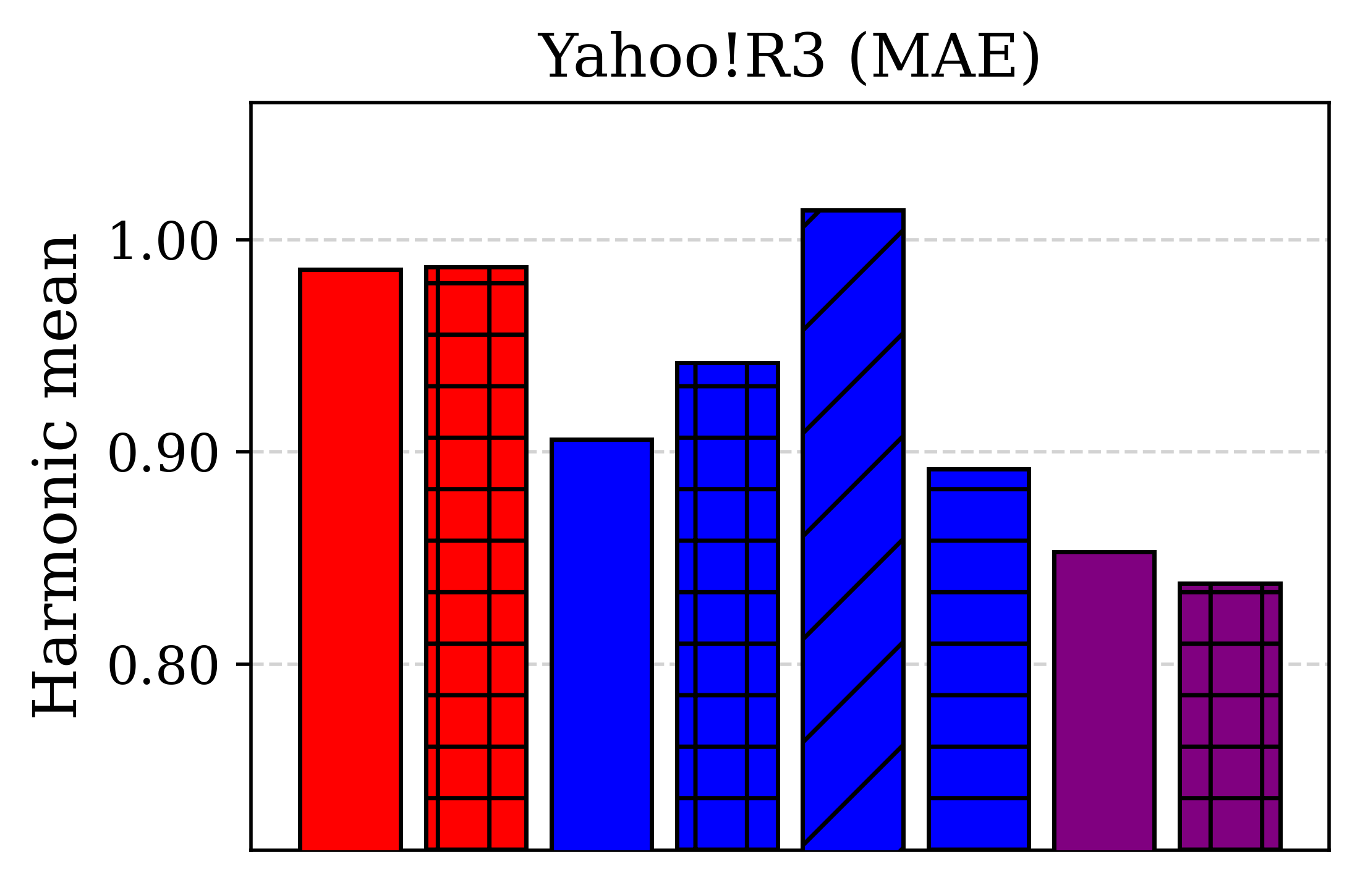}
\end{subfigure}
\begin{subfigure}[t]{0.27\linewidth}
    \includegraphics[width=\linewidth]{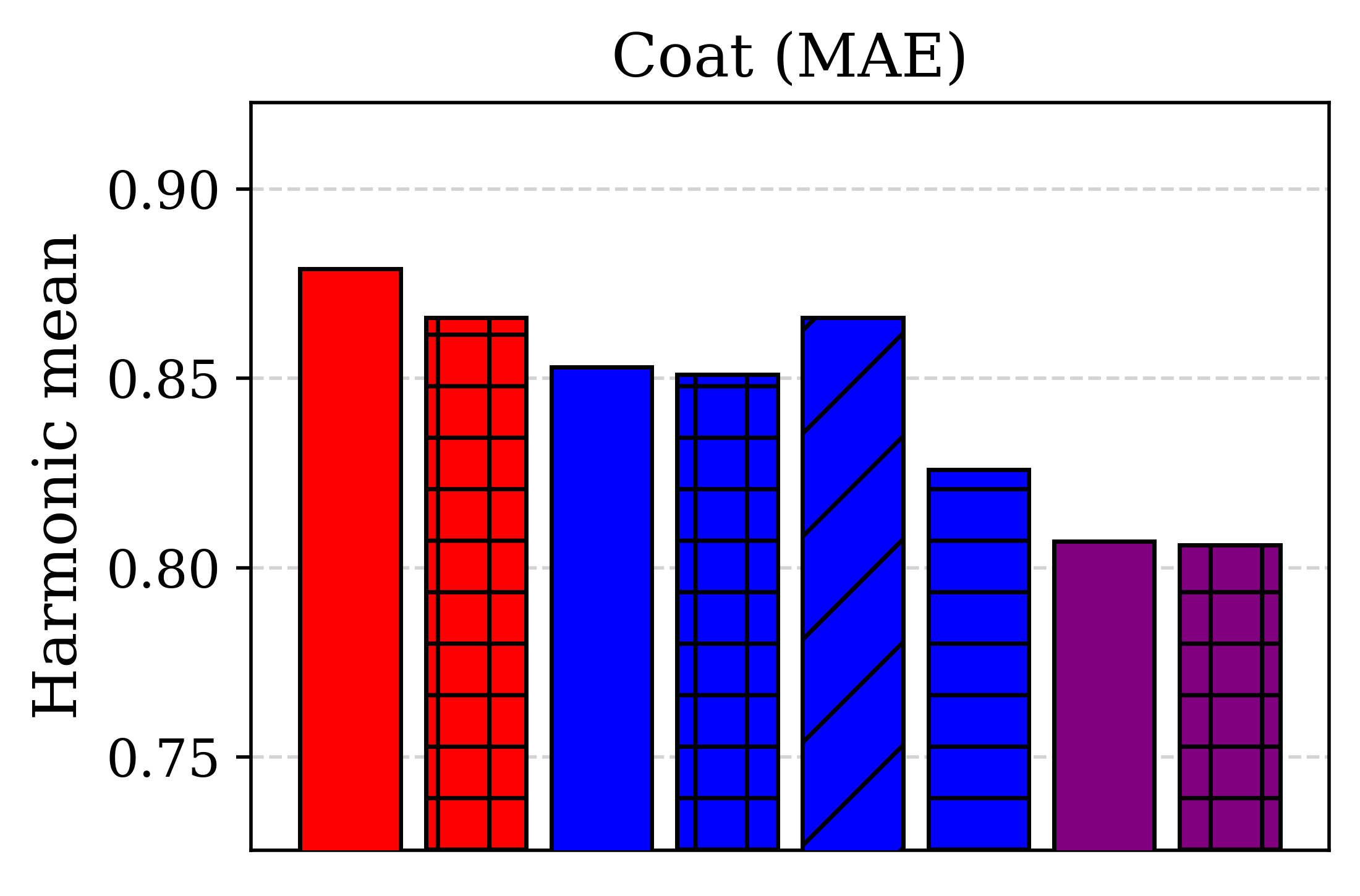}
\end{subfigure} 
\begin{subfigure}[t]{0.27\linewidth}
    \includegraphics[width=\linewidth]{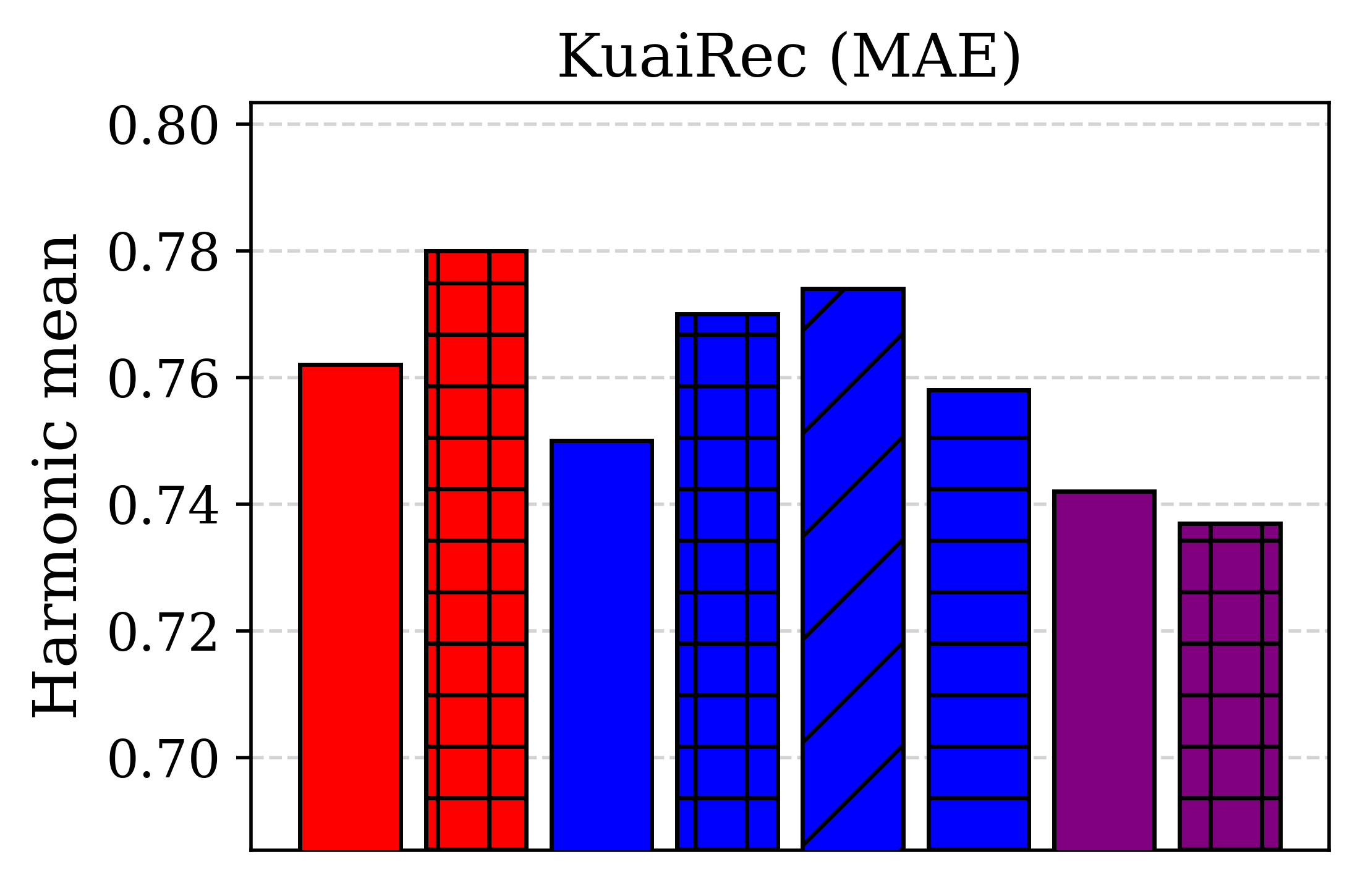}
\end{subfigure}
\caption{Harmonic mean of factual and counterfactual test results across three real-world datasets.
We perform a paired t-test with the best competitor at the 0.05 significance level. 
\proposed-Soft achieves statistically significant improvements in three out of six cases, and \proposed-Hard significantly outperforms in five out of six cases.
}
\label{fig:dual_h}
\vspace{-0.3cm}
\end{figure*}

%% file: Sections/091main_RCT.tex
\begin{table}[t!]
\caption{Benchmark counterfactual test (RCT). * denotes significance (paired t-test, $p < 0.05$) against the best baseline. 
}
\centering
\label{tab:main_table_b}
\renewcommand{\arraystretch}{0.8}
\resizebox{0.9\linewidth}{!}{
    \begin{tabular}{lcccc}
    \toprule
    \multirow{2}{*}{Method} & \multicolumn{2}{c}{Yahoo!R3} & \multicolumn{2}{c}{Coat} \\
    \cmidrule(lr){2-3} \cmidrule(lr){4-5}
      & MSE & MAE & MSE & MAE \\
    \midrule 
     Biased Teacher & 1.987 & 1.096 & 1.306 & 0.918 \\ 
     Standard Training  & 1.937 & 1.086 & 1.284 & 0.904 \\
     \midrule 
      ADV & 1.927 & 1.083 & 1.254 & 0.898 \\
      FADA & 1.499 & 0.956 & 1.162 & 0.838 \\
       FADA/IA  & 1.456 & 0.898 & 1.253 & 0.845 \\
     FreeMatch & 1.425 & 0.925 & 1.172 & 0.842\\ \midrule
    
      IPS  & 1.712 & 1.062 & 1.109 & 0.873 \\
      AST & 1.425 & 0.968	& 1.178	& 0.844 \\
      AT & 1.350 & 0.945 & 1.105 & 0.832 \\
        DAMF & 1.279 & 0.926 & 1.137 & 0.882 \\    
      MR & 1.349 & 0.892 & 1.106 & 0.786 \\
    TMRDR-CL  & 1.341 & 0.890 & 1.104 & 0.786 \\
    Stable-DR  & 1.327 & 0.887 & 1.088 & \textbf{0.778} \\
     Stable-MRDR  & 1.316 & 0.882 & 1.093 & \underline{0.779} \\ 
    DCE-TDR  & 1.319 & 1.015 & 1.091 & 0.832\\
    \midrule
      TIDE & 1.547 & 1.071 & 1.274 & 0.917 \\
      Zerosum & 1.474 & 0.894 & 1.228 & 0.896 \\
      ESAM & 1.480 & 0.897 & 1.129 & 0.853 \\
    InvCF & 1.398 & 1.021 & 1.194 & 0.897 \\  \midrule
      InterD  & \underline{1.278} & \underline{0.849} & \underline{1.052} & 0.814\\
     DebiasKD & 1.354 & 0.895 & 1.101 & 0.795\\ 
     \proposed-Soft & 1.086* & 0.787* & 1.006 & 0.782 \\
    \proposed-Hard & $\,\,$\textbf{0.987}$^{*}$ & $\,\,$\textbf{0.745}$^{*}$ & $\,\,$\textbf{0.986}$^{*}$ & \underline{0.779}\\ 
    \bottomrule
    \end{tabular}}
\end{table}

%% file: Sections/060Conclusion.tex

We propose \proposed, equipped with new dual distillation strategies, which consist of:
(1) reliability-filtered self-distillation, which iteratively refines the model to uncover preferences for unrated data. 
(2) confidence-penalized preference distillation, which supplements limited feedback using the biased teacher knowledge.
These two distillations are adaptively balanced based on the affinity of each potential user-item combination to the collected feedback.
Our comprehensive experiments show the effectiveness of \proposed in both factual and counterfactual test environments. 
Future work may explore the applicability of our method to other forms of feedback, such as implicit feedback.
\textcolor{red}{In addition, it is important to expand debiasing learning studies to less-studied real-world environments, such as federated learning settings \cite{wang2024bilateral, ali2024hidattack, ali2025privacy}.}

\textbf{Acknowledgments.}
This work was supported by Microsoft Research Asia and IITP (No.2022-00155958), the NRF funded by the Ministry of Education (NRF-2021R1A6A1A03045425), the IITP grants funded by the MSIT (IITP-2025-RS-2020-II201819, RS-2024-00457882), and 2025 High-Performance Computing Support Program.